\documentclass[sigconf]{acmart}

\AtBeginDocument{%
  }

\setcopyright{acmcopyright}

\copyrightyear{2023}
\acmYear{2023}
\setcopyright{rightsretained}
\acmConference[ASIA CCS '23]{ACM ASIA Conference on Computer and Communications Security}{July 10--14, 2023}{Melbourne, VIC, Australia}
\acmBooktitle{ACM ASIA Conference on Computer and Communications Security (ASIA CCS '23), July 10--14, 2023, Melbourne, VIC, Australia}
\acmDOI{10.1145/3579856.3582822}
\acmISBN{979-8-4007-0098-9/23/07}

\usepackage{booktabs} 
\usepackage{multirow}
\usepackage{algorithm}
\usepackage{algorithmic}

\begin{document}


\title{DHBE: Data-free Holistic Backdoor Erasing in Deep Neural Networks via Restricted Adversarial Distillation}


\author{Zhicong Yan}
\email{zhicongy@sjtu.edu.cn}
\orcid{0000-0003-3363-7713}
\affiliation{%
  \institution{Shanghai Jiao Tong University}
  \state{Shanghai}
  \country{China}
}

\author{Shenghong Li}
\authornote{Corresponding Author.}
\email{shli@sjtu.edu.cn}
\orcid{0000-0002-0767-2307}
\affiliation{%
  \institution{Shanghai Jiao Tong University}
  \state{Shanghai}
  \country{China}
}

\author{Ruijie Zhao}
\email{ruijiezhao@sjtu.edu.cn}
\orcid{0000-0001-6168-8687}
\affiliation{%
  \institution{Shanghai Jiao Tong University}
  \state{Shanghai}
  \country{China}
}

\author{Yuan Tian}
\email{ee_tianyuan@sjtu.edu.cn}
\orcid{0000-0001-6073-8582}
\affiliation{%
  \institution{Shanghai Jiao Tong University}
  \state{Shanghai}
  \country{China}
}

\author{Yuanyuan Zhao}
\email{yyzhao04@163.com}
\orcid{0000-0001-5645-363X}
\affiliation{%
  \institution{Hangzhou Normal University}
  \state{Zhejiang}
  \country{China}
}

\renewcommand{\shortauthors}{Yan et al.}

\begin{abstract}
  Backdoor attacks have emerged as an urgent threat to Deep Neural Networks (DNNs), where victim DNNs are furtively implanted with malicious neurons that could be triggered by the adversary. To defend against backdoor attacks, many works establish a staged pipeline to remove backdoors from victim DNNs: inspecting, locating, and erasing. However, in a scenario where a few clean data can be accessible, such pipeline is fragile and cannot erase backdoors completely without sacrificing model accuracy. To address this issue, in this paper, we propose a novel data-free holistic backdoor erasing (DHBE) framework. Instead of the staged pipeline, the DHBE treats the backdoor erasing task as a unified adversarial procedure, which seeks equilibrium between two different competing processes: distillation and backdoor regularization. In distillation, the backdoored DNN is distilled into a proxy model, transferring its knowledge about clean data, yet backdoors are simultaneously transferred. In backdoor regularization, the proxy model is holistically regularized to prevent from infecting any possible backdoor transferred from distillation. These two processes jointly proceed with data-free adversarial optimization until a clean, high-accuracy proxy model is obtained. With the novel adversarial design, our framework demonstrates its superiority in three aspects: 1) minimal detriment to model accuracy, 2) high tolerance for hyperparameters, and 3) no demand for clean data. Extensive experiments on various backdoor attacks and datasets are performed to verify the effectiveness of the proposed framework. Code is available at \url{https://github.com/yanzhicong/DHBE}
\end{abstract}




\maketitle

\section{Introduction}

Despite the unprecedented success of DNNs in various machine learning tasks, their reliability has been challenged by various kinds of attacks as a consequence of lacking interpretability of its decisions. Recently, the neural backdoor attacks (a.k.a. trojaning attacks) against DNNs gained extensive attention due to the great threat they brought \cite{li2020backdoor}. This kind of attack aims to construct a conditionally\textcolor{red}{-}triggered response between a specific input pattern and the target output desired by the adversary, so that the adversary could mislead the prediction of DNNs for subsequent malevolent activities.

\begin{figure}
    \centering
    \includegraphics[width=0.95\linewidth]{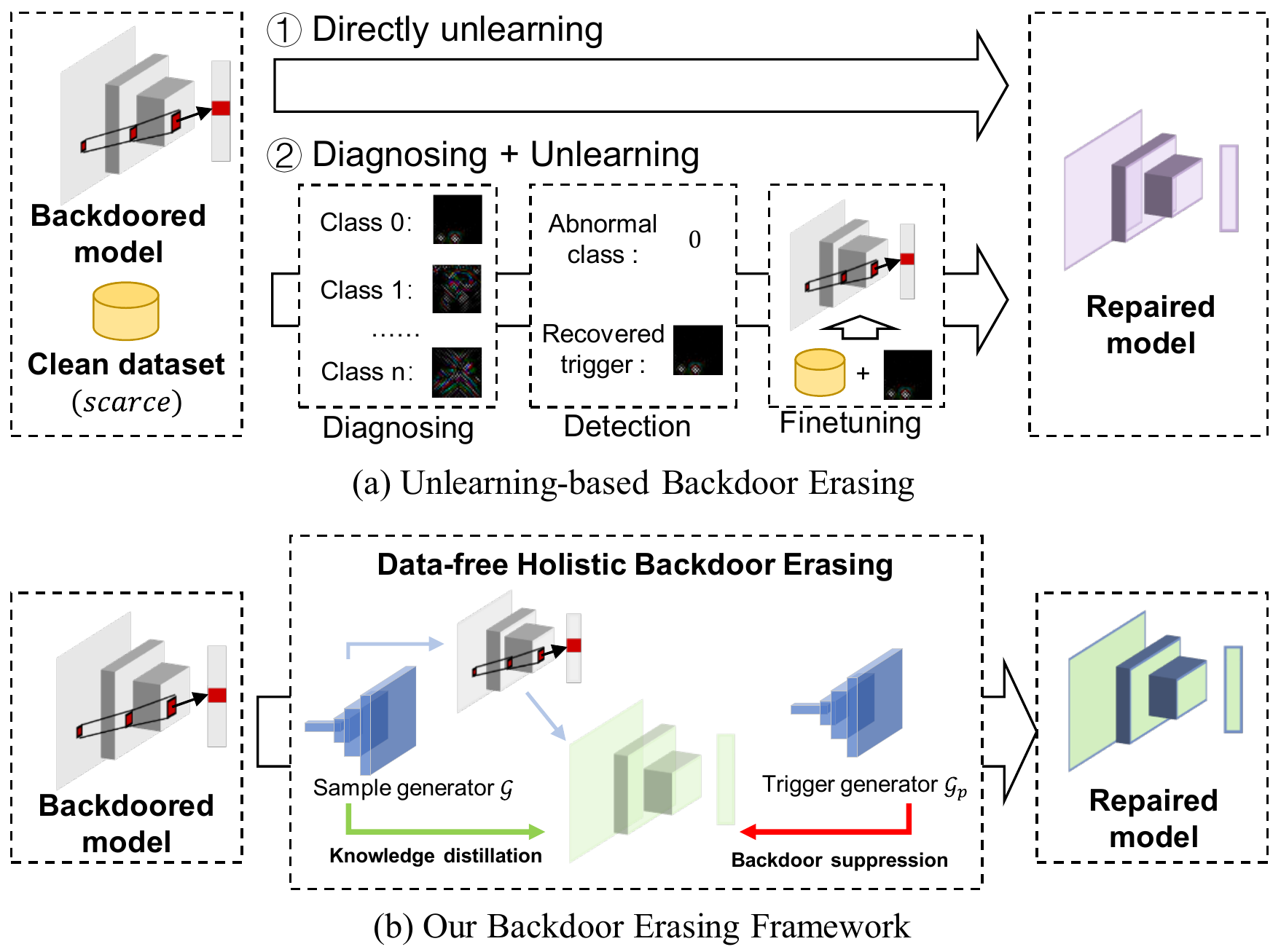}
    \vspace{-2mm}
    \caption{Comparison of unlearning-based backdoor erasing methods and our backdoor erasing framework. We propose to combine data-free knowledge distillation and backdoor suppression in a single framework, so that we can 1) minimize detriment to model's functionality, 2) erase backdoors without access to clean data.}
    \vspace{-2mm}
    \label{fig:illustrate}
\end{figure}

Previous backdoor defense methods focus on filtering training data to thwart attacks via data-poisoning \cite{chen2019detecting, tian2021self, hayase2021spectre, tran2018spectral}. However, as the supply chain of DNNs becomes increasingly complex, many corresponding defense methods that only focus on the training data are not effective anymore. Modern industrial DNN models are often delivered by third-party training platforms (such as Amazon SegeMaker, Huawei ModelArts, and Baidu PaddlePaddle, etc), the distributed model supply chain offers a new venue for other backdoor variants, such as poisoning the pre-trained models \cite{wang2020backdoor, yao2019latent, jia2022badencoder}, neuron hijacking \cite{liu2017trojaning}, and even code poisoning \cite{bagdasaryan2020blind}. To provide a comprehensive defense scheme to those backdoor attacks, the defense techniques that target the final delivered models, also known as \textit{model diagnosing and backdoor erasing}, provide more promising results and have drawn great research interest in recent years \cite{qiao2019defending, chen2019deepinspect, wang2019neural, li2021neural, liu2021removing}. Many model diagnosing methods \cite{chen2019deepinspect, xu2021detecting} have been developed to identify hidden backdoors, but the victim DNN models where backdoors are identified still need to be repaired by backdoor erasing methods. However, as the malicious neurons are entangled with normal neurons in attacked DNN models, the backdoor erasing task is quite challenging.

Existing backdoor erasing methods generally use the idea of unlearning where backdoored DNN models are finetuned \cite{wang2019neural}, pruned \cite{liu2018fine}, or distilled \cite{li2021neural} to ``forget" the backdoor. These methods seem to perform reasonably well in experimental settings, but are problematic in real-world defense scenarios. 
Two troubles are highlighted: 1) \textit{Hard to find an appropriate setting.}
During the unlearning process, the model not only forgets the injected backdoor but also compromises its functionality. 
As the intensity of unlearning increases (e.g. increasing learning rate, training epochs, etc.), the backdoor gets weaker, but the DNN model suffers from sharp performance degradation. 
However, different triggers have different resistance to unlearning, which may cause great trouble to the defender since it cannot measure the attack success rate of unknown triggers and can only use the most aggressive unlearning settings to minimize the probability of being attacked. 
Our evaluations show that Finepruning \cite{liu2018fine} and NAD \cite{li2021neural} have unacceptable model accuracy degradation (10\%) on Cifar10 dataset when all employed backdoors are wiped out.
2) \textit{Too depend on high quality dataset}. Unlearning methods are more destructive when the training data is scarce. For example, the NAD \cite{li2021neural} has about 5\% accuracy degradation holding 4\% of all clean data in the experimental scenario, but the accuracy degradation almost doubles when using 1\% clean data.
Someone may debate that the defender can collect more clean data to minimize the loss of model accuracy, but if the defender can collect enough clean data, why not train a new model from scratch?

To alleviate the negative effect of unlearning on model functionality, existing methods prefer to use a complex pipeline, where an optimization-based trigger reverse-engineering routine is performed first, then unlearning is executed to specifically mitigate the recovered trigger \cite{liu2017trojaning, wang2019neural, qiao2019defending, guo2019tabor, tian2022ean}. 
However, such a surgical pipeline increases the difficulty and complexity of backdoor erasing tasks, but only alleviates the aforementioned problems. 
What's more, these staged methods are easily compromised by newly-designed attacks with either trigger that is harder to be detected and recovered \cite{li2020invisible,8892628}, or composite trigger that contains multiple parts \cite{xie2019dba, lin2020composite, 9954194, tian2020self}.

\textbf{Our work: } In this work, we propose a novel \textit{Data-free Holistic Backdoor Erasing} (DHBE) framework. 
Instead of unlearning backdoors, the DHBE treats the backdoor erasing task as a unified adversarial procedure, where a clean model is obtained by combining data-free knowledge distillation and backdoor suppression (Fig. \ref{fig:illustrate}). 
Given a backdoored model $\mathcal{T}$, a clean model $\mathcal{S}$ is obtained by solving an optimization problem with two conflicting objectives:
1) \textit{functionality objective}, which minimizes the difference between the outputs of $\mathcal{T}$ and $\mathcal{S}$ over the entire input space, distilling the knowledge from $\mathcal{T}$ to $\mathcal{S}$, and 2) \textit{backdoor suppression objective}, which restricts the expected output change of $\mathcal{S}$ w.r.t. $\ell_1$-norm bounded trigger patterns. 
By jointly optimizing these two objectives based on data-free optimization strategies, the optimized model $\mathcal{S}^*$ finally reaches the desired equilibrium: it inherits the clean data accuracy of $\mathcal{T}$, and shows nearly no response to the trigger hidden in $\mathcal{T}$.

In DHBE, \textit{adversarial distillation} \cite{fang2019data} is employed to optimize the functionality objective. 
A sample generator $\mathcal{G}$ is designed to synthesize samples that have large discrepancies between the $\mathcal{T}$ and $\mathcal{S}$, then the model $\mathcal{S}$ is adversarially trained to minimize the discrepancy on generated samples. By dynamically updating $\mathcal{G}$, $\mathcal{S}$ is always convergent to $\mathcal{T}$ during the optimization,
keeping the accuracy of $\mathcal{S}$ always close to $\mathcal{T}$.
Along with distillation, we propose \textit{adversarial backdoor regularization} (ABR) to accomplish the backdoor suppression objective.
Since backdoor triggers are commonly designed to have a small $\ell_1$ norm to avoid being noticed, we define the surrounding space of input $x$ bounded by $\ell_1$ distance threshold as the trigger space of $x$, and the regularization term is proposed as the expected output changes of $\mathcal{S}$ when input $x$ changes within the trigger space. 
Another adversarial process is designed in ABR: 
A trigger generator $\mathcal{G}_p$ is optimized to synthesize the most sensitive trigger of $\mathcal{S}$ and $\mathcal{S}$ is adversarially optimized to erase the generated trigger.

Using DHBE, the defender could repair the backdoored model easily and effectively. First, the defender does not need to carefully adjust the hyperparameters. Since the DHBE is based on adversarial optimization rather than unlearning, the equilibrium state of $\mathcal{S}$ is so stable that a wide range of hyperparameters is feasible.
Second, the defender does not need to prepare clean datasets. Existing backdoor erasing methods require access to datasets to finetune the neural network, which may lead to privacy leakage;
Last, our framework offers a flexible plug-and-play service. For an application supported by multiple types of DNNs from Google, Tiktok, etc., deploying the DHBE is much more practical than deploying existing methods.


We conduct comprehensive evaluations involving four standard image datasets, attack methods with different backdoor injection mechanisms, and different sizes of patch triggers.
Our experimental results on the CIFAR10 show that the DHBE can wipe out all patch triggers with less than 1\% accuracy degradation, and its effectiveness is stable within a wide range of hyperparameters.
In contrast, the unlearning methods cost about 5\% model accuracy to erase a $2\times2$ square trigger and more to erase a $5\times5$ square trigger.
Moreover, we quantified the models' internal response to injected triggers after erasing, and observed that: 1) Unlearning-produced models still have a sensitive response to the trigger and can be attacked by amplified triggers (i.e. inject the same trigger multiple times to the input). 2) DHBE-produced models have nearly no sensitivity to the trigger, thus cannot be attacked by amplified triggers.
Our results indicate that the DHBE not only has much less negative impact on the model's functionality, but is also more secure than unlearning methods.

Our contributions can be summarized as follows:

\begin{itemize}
    \item We propose a novel Data-free Holistic Backdoor Erasing (DHBE) framework to cure the backdoored model and defend against backdoor attacks. To the best of our knowledge, this is the first backdoor erasing method that does not require extra training data.
    \item We propose adversarial backdoor regularization to prevent the student model in adversarial distillation from being infected by triggers with small $\ell_1$ norms, where a clean model is achieved by reaching the equilibrium state between these two adversarial processes.
    \item Extensive experiments are performed on standard image datasets to demonstrate the superiority of the proposed defense method against various backdoor attacks.
\end{itemize}

\section{Related Work}

\subsection{Backdoor Attacks}
The backdoor attacks need to temper the DNN model so that the tempered model is sensitive to a specific pattern that could be interpolated into any normal inputs \cite{gu2017badnets,yan2021dehib}. 
Attribute to the powerful perception ability of DNNs, various kinds of patterns (also known as ``triggers'' in backdoor attacks) had been shown to be capable of redirecting tempered models' predictions:
1) Small-size but intense triggers. Badnets \cite{gu2017badnets} had shown that the trigger could be as small as a single pixel or a small square of pixels in a large image, then this kind of trigger was investigated in most subsequent studies \cite{liu2017trojaning, xie2019dba, saha2020hidden, turner2019label}.
2) Large-size but invisible triggers.
Another line of studies employs a globally but subtle modification to original images as the backdoor triggers.
Barni \textit{et al.} \cite{barni2019new} superimposed the images with a global sinusoidal signal,
and Li \textit{et al.} \cite{li2020invisible} leveraged steganography where the least significant bits (LSBs) of pixels are substituted. 
These triggers were implanted with invisible modification of the entire image,
and demonstrated more robustness to the state-of-the-art defense methods.

\subsection{Backdoor Defenses}
Various backdoor defense methods were proposed, which could be categorized as follows according to their defense principle: poison data detection \cite{chen2019detecting, tran2018spectral, hayase2021spectre}, robust learning \cite{li2021anti, yan2021deep}, model diagnosing and backdoor erasing \cite{liu2018fine, liu2019abs, wang2019neural, qiao2019defending, guo2019tabor, li2021neural}, post-deployment input inspecting and augmentation \cite{doan2020februus, villarreal2020confoc}.

In this work, we focus on backdoor erasing, which tries to erase the backdoors hidden in backdoored models and recover a clean model that immune to backdoor triggers. 
Two main kinds of backdoor erasing methods have been investigated:


\textbf{(a) Blind Erasing Methods:}
In blind erasing methods, the defender directly employed finetuning, pruning \cite{liu2018fine}, distillation \cite{li2021neural} and so on, which could let the model unlearn the backdoor triggers. 
However, during the backdoor erasing process, useful visual clues were often erased along with the backdoor trigger, leading to an apparent descent of model accuracy.

\textbf{(b) Targeted Erasing Methods:}
In targeted methods, the defender first performed backdoor detection routines to recover the possible trigger patterns and the intention of the adversary, then tried to eliminate the recovered trigger by model retraining or pruning.
Neural Cleanse (NC) \cite{wang2019neural} proposed to generate a minimal $\ell_1$-norm trigger for each output class, and consider the class with the minimal $\ell_1$-norm trigger as the backdoor attacked class. 
Qiao \textit{et al.} \cite{qiao2019defending} improved NC by modeling the distribution of triggers with a staircase generative model, and Tabor \cite{guo2019tabor} designed a new objective function to find an optimal trigger for each class.

In this work, we propose a unified backdoor erasing framework, and design a data-free optimizing approach with stable analysis and much less damage to the model's accuracy on normal inputs.

\subsection{Data-free Knowledge Distillation}

Knowledge distillation had been proposed to transfer the performance from a cumbersome model to a small and efficient model \cite{hinton2015distilling}. However, it employed a data-driven approach in the distilling process, which was difficult to practice when the training data is scarce or not accessible. To overcome this issue, authors in \cite{chen2019data} and \cite{yin2020dreaming} tried to generate realistic synthesized images from the trained model that could be used for knowledge transfer. 
Fang \textit{et al.} \cite{fang2019data} designed a data-free adversarial distillation framework, where the training samples were crafted by a generator with the intention of maximizing the teacher-student discrepancy. Since the generator in \cite{fang2019data} took a long time to convergent, in \cite{fang2022up}, a meta-learning method was designed to accelerate the knowledge distillation process.

Data-free knowledge distillation is fast evolving in recent years, and readers are recommended to refer to \cite{liu2021data} for the latest advance. 
In this work, we choose \cite{fang2019data} as one component of the proposed framework to show their potential in the field of backdoor erasing tasks due to their power capability of transferring knowledge between different networks.



\section{Threat Model and Defense Settings}


Instead of adopting the previous popular settings \cite{wang2019neural, qiao2019defending, li2021neural, chen2019deepinspect} where a clean or synthetic dataset is available, we define a novel setting termed \textbf{Data-free Backdoor Erasing} 
where the defender holds a DNN model that has been diagnosed as being backdoor attacked with a high probability, but does not have access to clean data that could be used to finetune the model. 
Formally, the attacked DNN model is denoted as $\mathcal{T}:\mathcal{X} \mapsto \mathbb{R} ^{n_c}$, which takes image $x$ with size $H\times W \times C$ as inputs and output a class score vector $q \in \mathbb{R} ^{n_c}$. Specifically, the attacked DNN model predicts the operated images as the attacker-desired category $t$. $t$ is referred to as the target category while other categories are denoted as source categories $\{s_i\}$. 
The defender's goal is to transform the backdoored DNN model $\mathcal{T}$ into a clean DNN model $\mathcal{S}$ without access to any training data. 
However, what the defender knows is only the fact that an existing adversary could alter the model's predictions by making a specific operation $\mathcal{A}:\mathcal{X}\mapsto\mathcal{X}$ on the natural inputs $x \in \mathcal{X}$.

Although the only difference between our settings to previous works is the accessibility to clean datasets, we argue that this difference is critical in many real-world defense scenarios.
Since people are increasingly concerned about data privacy, building artificial intelligent systems without enough training datasets or without centralized datasets is more and more common in practice,
such as 1) federated learning, where the training data are dispersedly controlled by participants. 2) few-shot learning \cite{vinyals2016matching}, where each category in its test phase only has a few images that are far from enough for the unlearning process in backdoor erasing. and 3) zero-shot learning \cite{wang2018zero}, where no training samples but auxiliary information is provided.
In these scenarios, there is no access to the dataset that could be used for backdoor erasing, but they are potential victims of backdoor attacks.
Besides, for many popular tasks such as face recognition, optical character recognition, pedestrian detection, and so on, the most convenient practice is to download models from open-source projects, however, their training datasets may also be unavailable for privacy concerns.
What's more, under an adversarial environment, it is difficult to make sure all training data are clean, since more and more invisible backdoor attacks \cite{li2020invisible, li2021invisible, saha2020hidden} are proposed in recent years, however, using poisoned datasets for backdoor erasing is definitely infeasible \cite{li2021neural}.

We emphasize here the advantages of the proposed novel settings. \textbf{1) Practical}: Our defense frees the defender from dealing with training datasets. Many real-world scenarios such as federated learning, face recognition, and medical image diagnosing are inaccessible to data because of the prohibitive data-collection cost or privacy concerns. 
\textbf{2) Stable}: Previous methods use clean datasets or patched datasets to finetune the backdoored model, where their performance will severely decrease if the dataset is of poor quality (too few samples or imbalanced classes). We believe our work will have more general applications in the foreseeable future.


\section{Data-free Backdoor Erasing Framework}
In this section, we first provide an overview of the proposed framework, then the detailed components design is introduced in the next. Finally, we present the overall DHBE framework.

\subsection{Overview of Data-free Holistic Backdoor Erasing Framework}
In this paper, we propose a novel data-free backdoor erasing framework, which could repair the backdoored model in a single procedure without access to clean data. In the proposed framework, the backdoored model $\mathcal{T}$ is distilled into a clean model $\mathcal{S}$, where a generalized objective function is formulated as:
\begin{equation}
\mathcal{S} = \underset{\mathcal{S}}{\mathrm{argmin}} \; \mathcal{L}(\mathcal{T},\mathcal{S}) = \underset{\mathcal{S}}{\mathrm{argmin}} \; \mathcal{D}(\mathcal{T}, \mathcal{S}) + \lambda \mathcal{R}(\mathcal{S}).
\label{eq:obj_function}
\end{equation}

The first term $\mathcal{D}(\mathcal{T}, \mathcal{S})$ is designed to measure the discrepancy between outputs of $\mathcal{T}$ and $\mathcal{S}$. Minimizing this discrepancy term is equivalent to transferring the backdoored model's knowledge to the student model. The second term $\mathcal{R}(\mathcal{S})$ is a regularization term that tries to mitigate the possible backdoors in the student model $\mathcal{S}$. By jointly minimizing these two terms using data-free adversarial optimization, we hope to obtain a student model that inherits the teacher's performance on clean data, but without backdoor reactions.

In DHBE, we design two coupled adversarial processes to simultaneously optimize these two terms ($\mathcal{D}$ and $\mathcal{S}$) in the objective function (Eq.\ref{eq:obj_function}) over the entire data domain, which we denote as \textit{adversarial distillation} (described in subsection \ref{lbl:adversarial_distillation}) and \textit{adversarial backdoor regularization} (described in subsection \ref{lbl:adversarial_backdoor_regularization}) respectively. 
Finally, we summarize the detailed training process of DHBE in subsection \ref{lbl:overall_dhbe_framework}.

\begin{figure*}
    \centering
    \includegraphics[width=0.8\linewidth]{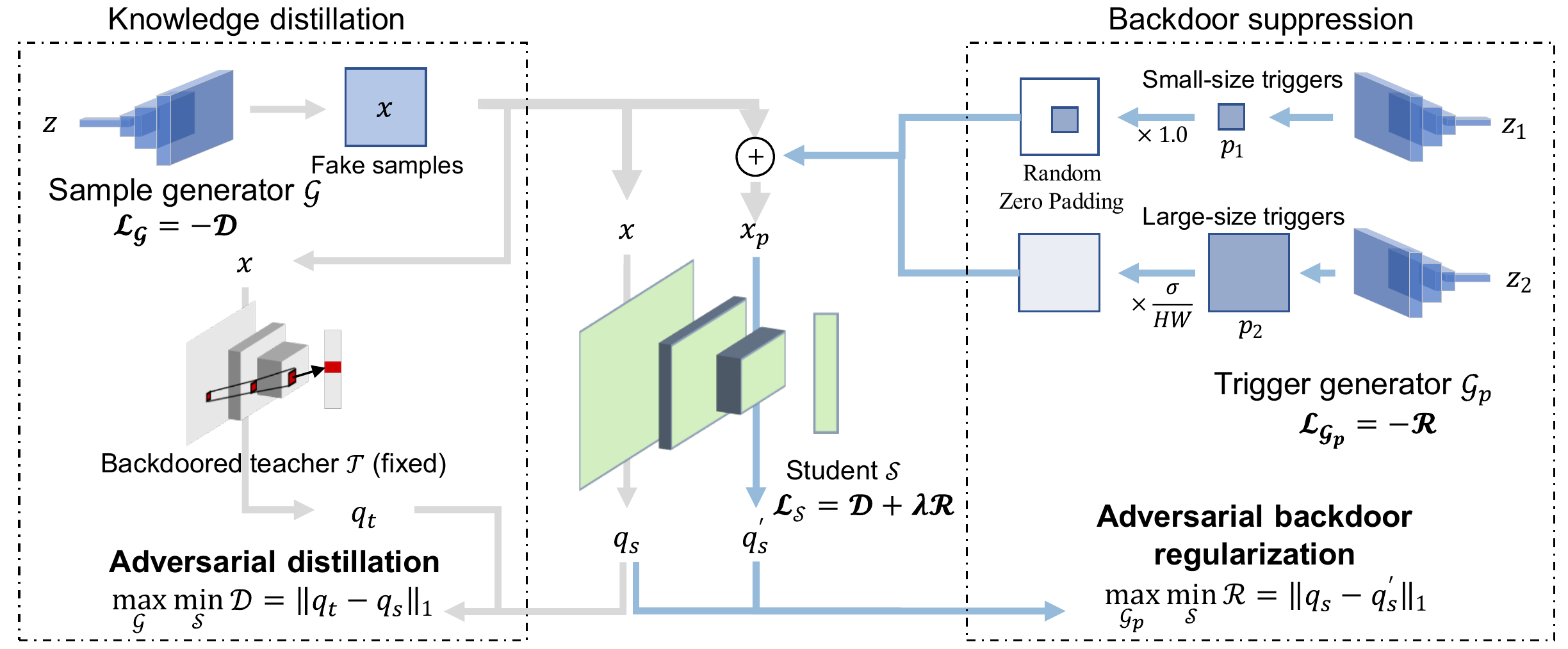}
    \vspace{-2mm}
    \caption{Overview of the Data-free Holistic Backdoor Erasing framework. Two adversarial processes are illustrated. 
    Adversarial distillation is designed to transfer the backdoored model's reactions to the student model, including both the normal reactions and backdoor reactions.
    Adversarial backdoor regularization is designed to suppress the backdoor triggers in the student model. 
    The student model tries to minimize a combined loss function with two adversarial objectives.}
    \vspace{-2mm}
    \label{fig:framework}
\end{figure*}

\subsection{Adversarial Distillation (AD)}
\label{lbl:adversarial_distillation}

We employ data-free adversarial distillation to transfer the teacher's knowledge to the student, which does not suffer from accuracy degradation caused by incomplete datasets. Intuitively, as the common practice of knowledge distillation, the original backdoored model is fixed and employed as the teacher model $\mathcal{T}$, then a student model $\mathcal{S}$ is optimized to mimic the output of the teacher model. Instead of using fixed training data as inputs, we design a hard sample generator $\mathcal{G}:\mathbb{R}^n \mapsto \mathcal{X}$ to dynamically generate training samples that cause large discrepancies between $\mathcal{T}$ and $\mathcal{S}$ during the training process. In the meantime, the student model $\mathcal{S}$ is adversarially trained to minimize the discrepancy on the generated samples. In our framework, the discrepancy between $\mathcal{T}$ and $\mathcal{S}$ is designed as the expected Mean Absolute Error (MAE) of model's pre-softmax outputs over randomly generated inputs following \cite{fang2019data}:
\begin{equation}
    \mathcal{D}(\mathcal{T},\mathcal{S};\mathcal{G})=\mathbb{E}_{z\sim p_z(z)} \bigg[  \Big\lVert \mathcal{T}(\mathcal{G}(z)) - \mathcal{S} (\mathcal{G}(z)) \Big\rVert_1
    \bigg],
\end{equation}
where $z$ is randomly sampled from normal distribution. While the teacher $\mathcal{T}$ is fixed, the sample generator $\mathcal{G}$ and the student $\mathcal{S}$ are iteratively trained to maximize and minimize the objective function respectively. Once the student catches up with the teacher over currently generated samples, the sample generator will move forward to the next confusing space. Finally, the student approaches the teacher over the entire input space, and inherits the teacher's accuracy on clean inputs.

If there is no regularization term to restrict the student model, the student model will learn the backdoor reactions from the teacher as well, since adversarial distillation forces it to comply with the teacher's behavior over the entire input space. In the next subsection, we characterize the backdoor reactions from a geometrical perspective, and propose a regularization term that could comprehensively erase all the backdoor reactions.

\subsection{Adversarial Backdoor Regularization (ABR)}
\label{lbl:adversarial_backdoor_regularization}

In this subsection, we describe the common characteristic of backdoor reactions in terms of how the output of the backdoored model changes when traveling through the input space. By characterizing the backdoor reactions without clean data distribution priori, we are able to distinguish the backdoor reactions from normal reactions in the backdoored model. Finally, we propose a backdoor regularization term and adversarially optimize it to erase the backdoor reactions.

\begin{table}[b]
    \caption{The expected $\ell_1$ EM distance between $\rho_s$ and $\rho_{s'}$ in different kinds of backdoor attacks. Here, $H$,$W$ and $C$ are height, width and channels of input images, $h$ and $w$ are height and width of square trigger patterns.}
    \centering
    \begin{tabular}{c|c}
    \toprule
      Trigger Type   & Expected $\ell_1$ EM Distance \\
    \hline
    Pixel \cite{gu2017badnets} & $\leq 1$  \\
    Square \cite{liu2017trojaning} &  $\leq h\times w \times C$  \\
    Watermark / Refool \cite{liu2020reflection} & $ \leq Opacity \times H \times W \times C$ \\
    SIG \cite{barni2019new} &  $ \leq \Delta/\sqrt{2}  \times H \times W \times C $  \\
    Steganograph \cite{li2020invisible} (2 bits) &  $ \leq 3/255 \times H \times W \times C$  \\
    \bottomrule
    \end{tabular}
    \label{tab:common_trigger_distances}
\end{table}

\begin{table}[b]
    \caption{We approximate the $\ell_1$ EM distance between all the class pairs within each dataset using \cite{kolouri2019generalized} and list the mean and minimal distances and the closest class pairs.}
    \centering
    \begin{tabular}{c|l|l}
    \toprule
       Dataset (Nb. dims)  & Closest Class Pair : Dist. & Mean Dist. \\
    \hline
    MNIST (784) & ``7'' $\rightarrow$ ``9'' : 168.88 &  495.26  \\
    \hline
    CIFAR10 (3072) & ``cat'' $\rightarrow$ ``dog'' : 282.76 & 869.83 \\
    \hline
    CIFAR100 (3072) & ``leopard'' $\rightarrow$ ``tiger'' : 292.16 & 1060.27  \\
    \bottomrule
    \end{tabular}
    \label{tab:closest_class_pairs}
\end{table}


\subsubsection{Characteristics of Backdoor Reactions}
Currently, backdoor attacks try to design the operation on clean images as subtle as possible to avoid being noticed, which means the backdoored model is forced to be extremely sensitive to triggers with small $\ell_1$ distances designed by the adversary. This rule is implicitly abided by various backdoor attacks: attacks with small-size triggers try to modify as small number of pixels as possible, and attacks with large-size triggers try to change the value of pixels as small as possible. More precisely, we formulate this characteristic as follows:

\begin{proposition}
Let $\rho_s$, $\rho_t \in Prob(\mathcal{X})$ be the probability measures of source class and target class, and $\rho_{s'}$ be the probability measure of backdoored inputs. Then the Earth-Mover (EM) distance (a.k.a. Wasserstein distance) between $\rho_s$ and $\rho_{s'}$ is far smaller than that between $\rho_s$ and $\rho_t$:
\vspace{-4mm}
\end{proposition}
\begin{equation}
    \mathbb{W}(\rho_s, \rho_{s'}) \ll \mathbb{W} (\rho_s, \rho_t),
\end{equation}
where the EM distance \cite{arjovsky2017wasserstein} is defined as the minimum cost of all the possible transport plans ($\Pi(\rho_s, \rho_t)$) from one probability measure to another under $\ell_1$ cost function:
\begin{equation}
    \mathbb{W}(\rho_s, \rho_t) = \inf_{\gamma \in \Pi(\rho_s, \rho_t)} {\mathbb{E}_{(x,y) \sim \gamma} \big[\rVert x - y \lVert_1 \big]}.
\end{equation}

The EM distance between $\rho_s$ and $\rho_t$ is hard to tackle, we approximate the distances within standard image datasets using Sliced Wasserstein Distances \cite{kolouri2019generalized}, and list the closest class pairs in Table \ref{tab:closest_class_pairs}. The EM distance between $\rho_s$ and $\rho_{s'}$ can be calculated as the sum of the total pixel changes by the backdoor operation. We list the distance approximation method of common triggers in Table \ref{tab:common_trigger_distances}. It can be seen that $\mathbb{W}(\rho_s, \rho_{s'})$ is commonly designed to be far smaller than $\mathbb{W} (\rho_s, \rho_t)$ to keep stealthy. For example, a square trigger for CIFAR10 ($32\times32$) may have up to the size of $10\times10$ to reach the distance of the closest class pair in CIFAR10, which is too obvious that can be easily discovered.

We note that assuming $\mathbb{W}(\rho_s,\rho_t)$ is large may not hold in some fine-grained classification tasks, such as face recognition, where the distance between two classes is relatively smaller than in other tasks. In the experiment section, we provide experimental results on fine-grained dataset (classification on a subset of classes in VGGFace2).


\subsubsection{Backdoor Regularization}
To get rid of the backdoor reactions, the student model should be smooth and robust to perturbations with small $\ell_1$ distances. To this end, we design a regularization term that forces the student model to predict the same result for any input $x$ and samples in its surrounding space $x' \in \Sigma(x, \varepsilon)$:
\begin{equation}
    \mathcal{R}(\mathcal{S}) = \mathbb{E}_{x\sim \mathcal{X}, x' \sim \Sigma(x, \varepsilon)} \Big[ \big\lVert  \mathcal{S}(x) - \mathcal{S}(x')  \big\rVert_1 \Big],
\end{equation}
where $\Sigma(x, \varepsilon)$ is a surrounding space of $x$ bounded by $\ell_1$ distance:
\begin{equation}
    \Sigma(x, \varepsilon) = \{ x' |  \lVert x - x' \rVert _1 < \varepsilon  \}.
\end{equation}

\subsubsection{Adversarial Optimization}

Directly optimizing the regularization term is equivalent to training the student model on the production of the sample space $\mathcal{X}$ and the corresponding trigger space $\Sigma$, which will fail since the production of two spaces is too large to be optimized. Thus, we propose another coupled adversarial process to optimize this term: On the one hand, the input $x$ is sampled from the sample generator $\mathcal{G}$ used in adversarial distillation. On the other hand, a trigger generator $\mathcal{G}_p : \mathbb{R}^n \mapsto \mathbb{R}^{h\times w\times C}$ is designed to generate triggers with size $h\times w$ that are most sensitive to $\mathcal{S}$, and enforce the following regularization term to mitigate the generated triggers:
\begin{equation}
   \mathcal{R}(\mathcal{S};\mathcal{G},\mathcal{G}_p) = \mathbb{E}_{x \sim \mathcal{G}, z\sim p_z(z)}
\bigg[
    \Big\lVert 
        \mathcal{S}(x) 
        - \mathcal{S} \big(x+\mathcal{G}_p(z) \big) \Big\rVert _1
\bigg].
\label{eq:regularization_term}
\end{equation}

To constraint the $\ell_1$-norm of the trigger generator's output below a given threshold $\varepsilon$, we design a trigger generator with size-fixed outputs ($h \times w$), and multiple the outputs with a constant scalar $s \in (0, 1]$, randomly pad to the same size of $x$ with zeros before finally adding up to the fake sample $x$. We use $Tanh$ as the output activation layer of $\mathcal{G}_p$, so the max $\ell_1$-norm of $\mathcal{G}_p$'s outputs is $h \times w \times s \times C$ ($C$ is image channels).

\begin{algorithm}[t]
\small
  \caption{Data-Free Blind Backdoor Erasing}
  \label{alg:training}
\begin{algorithmic}[1]
  \STATE {\bfseries Input:} A backdoored teacher model $\mathcal{T}(\cdot,\theta_t)$, batch size $BS$, $\lambda$, $\alpha_{\text{tv}}$, learning rates $\alpha_s$, $\alpha_g$, $\alpha_{gp}$.
  \STATE {\bfseries Output:} A clean student model $\mathcal{S}(\cdot,\theta_s)$.
  \STATE Initialize the student model's weights $\theta_s$ with $\theta_t$.
  \STATE Randomly initialize the sample generator $\mathcal{G}(\cdot,\theta_g)$ and the trigger generator $\mathcal{G}_p(\cdot,\theta_{gp})$.
  \FOR{number of training iterations}
        \FOR{$k$ steps}
            \STATE Randomly generate $BS$ samples $\{x_i\}$ and $BS$ triggers $\{p_i\}$ with $\mathcal{G}$ and $\mathcal{G}_p$;
            \STATE Randomly padding $\{p_i\}$ to the same size of $\{x_i\}$ with zeros; 
            \STATE $\mathcal{L}_s = 1/BS \sum_{i} ( \lVert \mathcal{T}(x_i) - \mathcal{S}(x_i) \rVert_1 + \lambda  \lVert \mathcal{S} (x_i) - \mathcal{S}(x_i + p_i) \rVert_1$);
            \STATE Update $\theta_s \leftarrow \theta_s - \alpha_s \nabla_{\theta_s} \mathcal{L}_s$;
        \ENDFOR
        \STATE Randomly generate $BS$ samples $\{x_i\}$ with $\mathcal{G}$;
        \STATE $\mathcal{L}_g = - 1/BS \sum_{i} ( \lVert \mathcal{T}(x_i) - \mathcal{S}(x_i) \rVert_1)$;
        \STATE Update $\theta_g \leftarrow \theta_g - \alpha_g \nabla_{\theta_g} \mathcal{L}_g$;
        \STATE Randomly generate $BS$ samples $\{x_i\}$ and $BS$ triggers $\{p_i\}$ with $\mathcal{G}$ and $\mathcal{G}_p$;
        \STATE Randomly padding $\{p_i\}$ to the same size of $\{x_i\}$ with zeros; 
        \STATE $\mathcal{L}_{gp} = - 1/BS \sum_{i} \lVert \mathcal{S} (x_i) - \mathcal{S}(x_i + p_i) \rVert_1$;
        \STATE Update $\theta_{gp} \leftarrow \theta_{gp} - \alpha_{gp} \nabla_{\theta_{gp}} \mathcal{L}_{gp}$;
  \ENDFOR
\end{algorithmic}
\end{algorithm}

\subsection{Overall DHBE Framework}
\label{lbl:overall_dhbe_framework}

The overall DHBE framework is illustrated in Fig. \ref{fig:framework}.
, where the student is adversarially optimized with two generators. We summarize the adversarial version of objective function (Eq. \ref{eq:obj_function}) as follows:
\begin{equation}
 \underset{\mathcal{G},\mathcal{G}_p}{\mathrm{max}} \; \underset{\mathcal{S}}{\mathrm{min}} \; \mathcal{L}(\mathcal{T},\mathcal{S};\mathcal{G},\mathcal{G}_p) = \underset{\mathcal{G},\mathcal{G}_p}{\mathrm{max}} \; \underset{\mathcal{S}}{\mathrm{min}} \; \mathcal{D}(\mathcal{T}, \mathcal{S}; \mathcal{G}) + \lambda \mathcal{R}(\mathcal{S}; \mathcal{G}, \mathcal{G}_p),
\label{eq:adversarial_obj_function}
\end{equation}
To accelerate the training process, we initialize the student model with the backdoored teacher model, then sequentially train the student and the generators like the training process of GANs \cite{goodfellow2014generative}.
In each iteration, we first update the student model $\mathcal{S}$ $k$ times (same as \cite{fang2019data}, we set $k=3$) to minimize the loss function that combines the discrepancy term and the regularization term, then the sample generator $\mathcal{G}$ and the trigger generator $\mathcal{G}_p$ are updated to maximize the discrepancy term and the regularization term respectively.
This iteration step is repeated thousands of times and the equilibrium state is achieved by learning rate annealing. We summarize the training process of the proposed DHBE framework in Alg. \ref{alg:training}.

\section{Experiments}

In this section, we first describe our experiment settings, then we compare the effectiveness of DHBE with both targeted and blind erasing methods on several well-known backdoor attacks.
Finally, we provide comprehensive ablation analyses.

\subsection{Experimental Settings}

\subsubsection{Evaluation Datasets}

Four standard image datasets are employed to evaluate the proposed framework, including three standard image datasets that are commonly used in various tasks: CIFAR10, CIFAR100 \cite{CIFAR}, Mini-Imagenet \cite{dhillon2019baseline}. Besides, we employ another face recognition dataset, VGGFace2 \cite{cao2018vggface2}, to show that our framework could perform well on fine-grained tasks.

\subsubsection{Configurations for Backdoor Attacks}
We employ three backdoor attacks with different backdoor injecting mechanism: Data-poisoning (Badnets\cite{gu2017badnets}, Clean-label \cite{turner2019label}) and Neuron hijacking (Trojaning \cite{liu2017trojaning}). We use 3 different size of triggers for each attack method.
For a fair comparison, we reimplement these attacks, and create backdoored models using the same Resnet-18 architecture \cite{he2016deep} provided by PyTorch \cite{paszke2017automatic}. 
As a common practice for training small datasets with Resnet-18, the $conv1$ layer ($kernal\; size=7,stride=2$) is replaced by $conv$ ($kernal\;size=3,stride=1$) and the first $Pooling$ layer is canceled to deal with inputs of size $32 \times 32$ (i.e. CIFAR10 and CIFAR100 in our experiments). For inputs of size $64 \times 64$ (i.e. Mini-Imagenet and VGGFace2 in our experiments), the $conv1$ layer is replaced by $conv$ ($kernal\;size=5,stride=2$).

\begin{table*}[]
    \centering
    \caption{Comparison results of DHBE to other blind erasing methods on CIFAR10 dataset against different backdoor attacks and different size of triggers. Numbers are displayed as percentages. }
    
	\begin{tabular}{c|c|cc|cc|cc|cc|cc|cc}
	\toprule
		\multirow{3}{*}{  \shortstack{Attack\\Methods}} &
		\multirow{3}{*}{  \shortstack{Trigger\\Size}} &  
			\multicolumn{2}{c|}{\multirow{2}{*}{\shortstack{Backdoored\\$t$='truck'}}} & 
				\multicolumn{2}{c|}{Finetuning} &
					\multicolumn{2}{c|}{Finepruning \cite{liu2018fine}} &
						\multicolumn{2}{c|}{MCR \cite{zhao2019bridging}} &
							\multicolumn{2}{c|}{NAD \cite{li2021neural}} &
								\multicolumn{2}{c}{DHBE}  \\
		&  &  
			 & &
				\multicolumn{2}{c|}{$N_{clean}=2000$} &
					\multicolumn{2}{c|}{{$N_{clean}=2000$}} &
						\multicolumn{2}{c|}{{$N_{clean}=2000$}} &
							\multicolumn{2}{c|}{{$N_{clean}=2000$}} &
								\multicolumn{2}{c}{No data required}  \\
		 & &  ACC & ASR &  ACC & ASR &  ACC & ASR &  ACC & ASR &  ACC & ASR  &  ACC & ASR  \\
	\hline

	\multirow{3}{*}{Badnets \cite{gu2017badnets}} & $2\times 2$ & 
 94.75 & 100.00  &  90.69 & 2.64  &  91.91 & 12.92  &  88.70 & 0.18  &  90.75 & 0.38  &  \textbf{94.05} & \textbf{0.02}  \\
	\cline{2-14}
	& $3\times 3$ &  
 95.02 & 100.00  &  90.67 & 2.97  &  91.48 & 6.74  &  89.17 & 1.92  &  90.87 & 7.08  &  \textbf{94.46} & \textbf{0.12}  \\
	\cline{2-14}
	& $5\times 5$ & 
 95.11 & 100.00  &  91.47 & 1.10  &  92.30 & 69.07  &  87.96 & 1.78  &  90.41 & 3.26  &  \textbf{94.24} & \textbf{0.00}  \\

	\hline

	\multirow{3}{*}{Clean-label \cite{turner2019label}} & $2\times 2$ & 
 94.85 & 46.32  &  90.51 & 1.71  &  92.33 & 6.51  &  93.49 & 1.75  &  90.28 & 1.22  &  \textbf{94.19} & \textbf{0.14}  \\
	\cline{2-14}
	& $3\times 3$ &  
 95.04 & 93.45  &  90.48 & 6.61  &  90.99 & 38.95  &  93.39 & 3.79  &  90.21 & 9.98  &  \textbf{94.11} & \textbf{1.17}  \\
	\cline{2-14}
	& $5\times 5$ & 
 95.25 & 100.00  &  91.08 & 96.41  &  91.79 & 67.44  &  93.74 & 14.01  &  90.69 & 8.57  &  \textbf{94.45} & \textbf{4.48}  \\

    \hline

	\multirow{3}{*}{Trojaning \cite{liu2017trojaning}} & $2\times 2$ & 
 94.56 & 56.20  &  89.75 & 0.09  &  91.98 & 0.90  &  88.65 & 0.16  &  91.24 & \textbf{0.09}  &  \textbf{93.96} & 4.39  \\
	\cline{2-14}
	& $3\times 3$ &  
 95.00 & 89.91  &  90.43 & 0.17  &  91.83 & 4.37  &  88.90 & \textbf{0.08}  &  90.49 & 1.25  &  \textbf{94.40} & 2.88  \\
	\cline{2-14}
	& $5\times 5$ & 
 94.93 & 99.91  &  90.05 & 17.05  &  92.03 & 68.82  &  87.79 & 0.84  &  90.67 & \textbf{0.55}  &  \textbf{94.56} & 1.13  \\

	\hline

	\multicolumn{2}{c|}{Mean ACC/ASR} & 
 94.95 & 87.31  &  90.57 & 14.31  &  91.85 & 30.63  &  90.20 & 2.72  &  90.62 & 3.60  &  \textbf{94.27} & \textbf{1.59}  \\
	\bottomrule

	\end{tabular}

    \label{tab:results_cifar10_small_trigger_blinded}
\end{table*}

\begin{figure*}[htb]
    \centering
    \begin{tabular}{ccc}
    \includegraphics[width=0.30\textwidth]{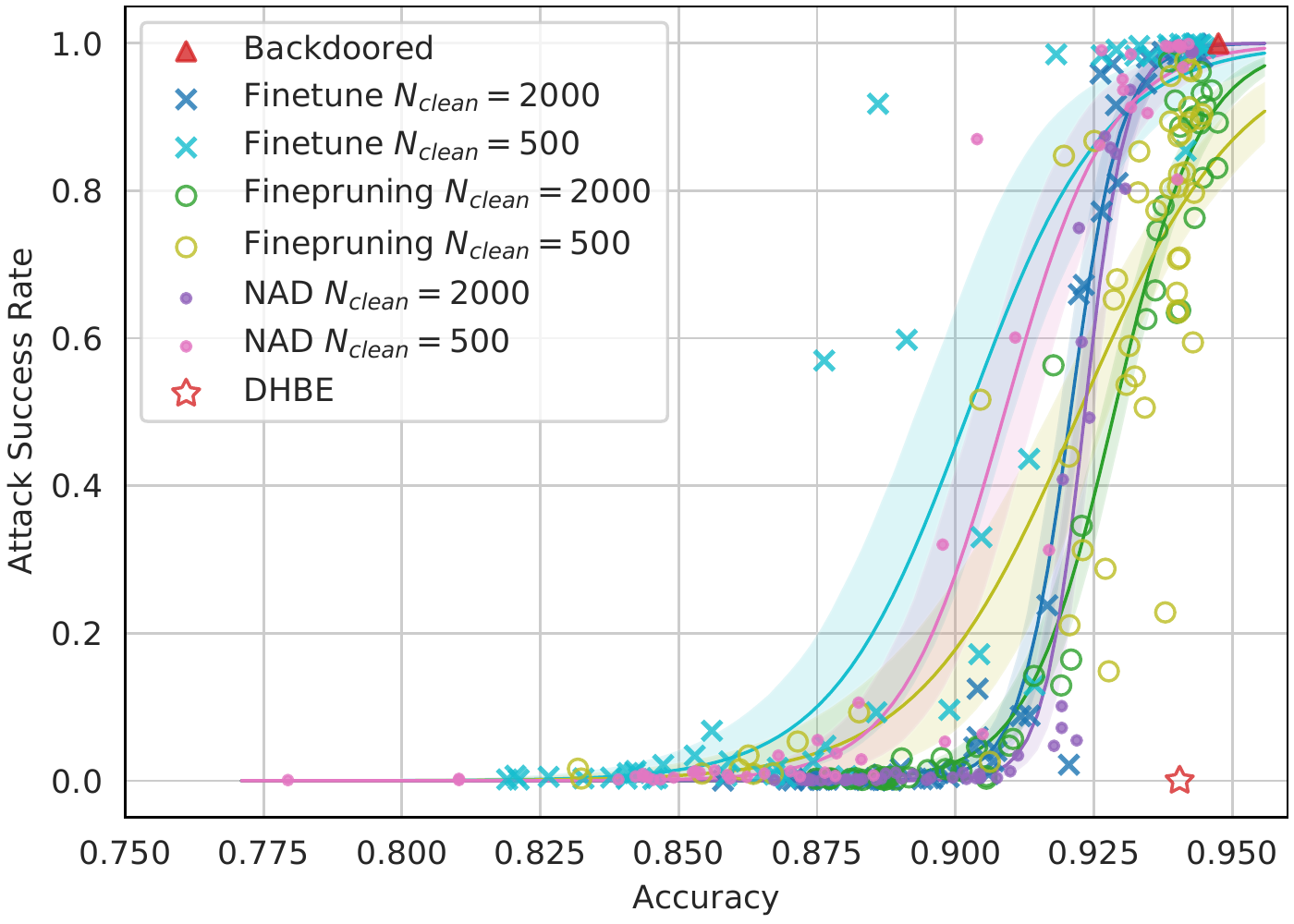}&
    \includegraphics[width=0.30\textwidth]{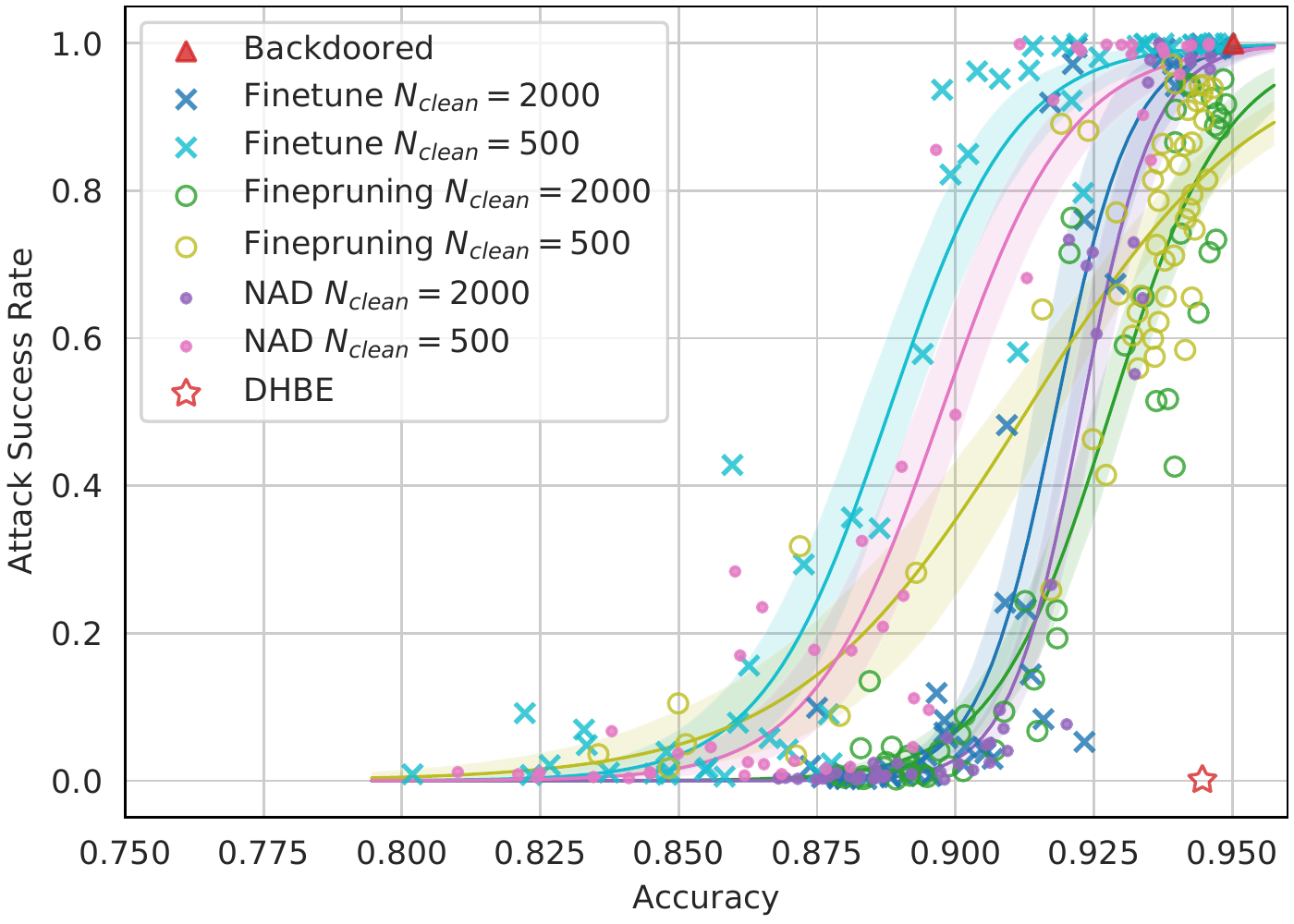}&
    \includegraphics[width=0.30\textwidth]{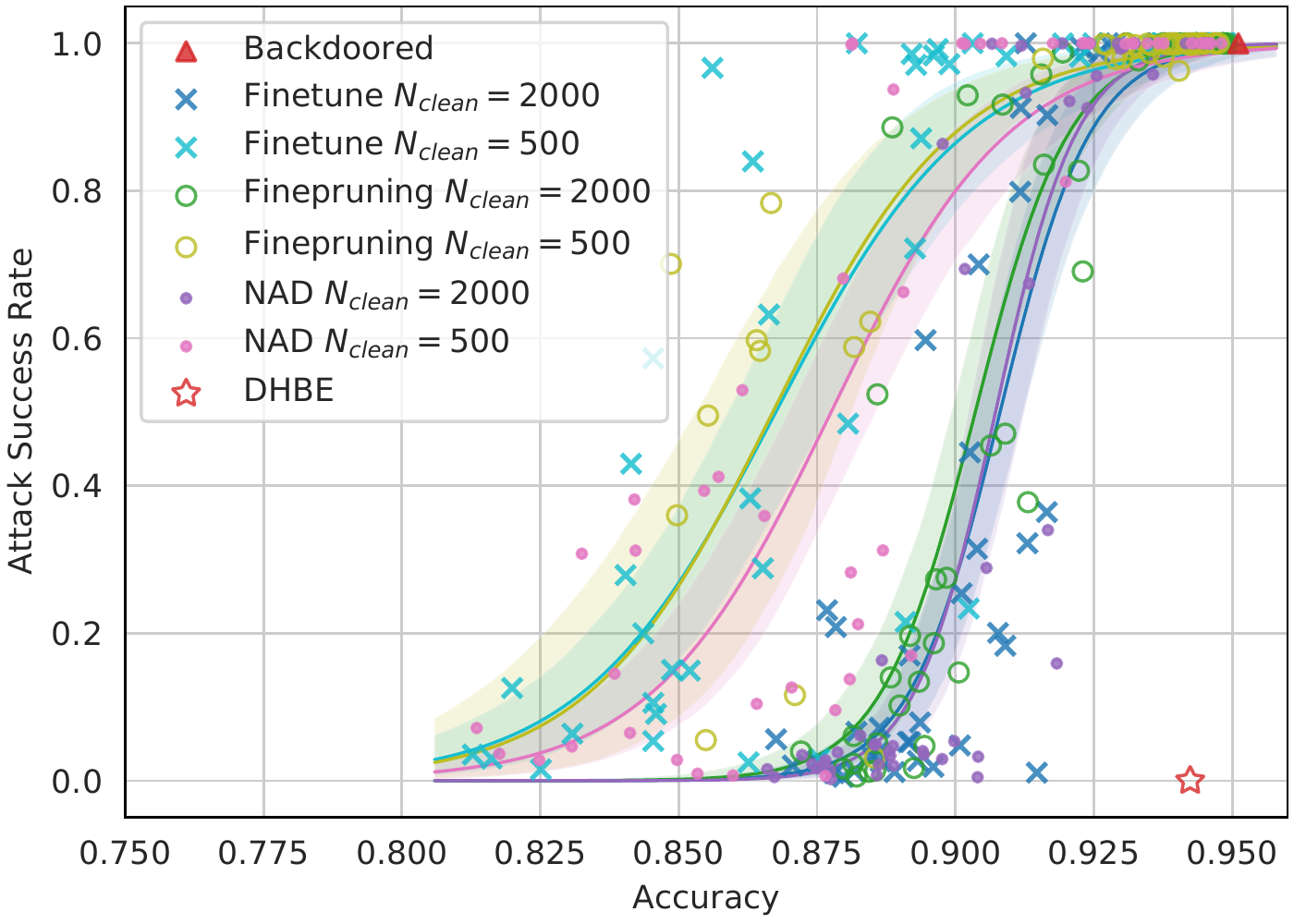}\\
    (a) Badnets ($2\times 2$)&
    (b) Badnets ($3\times 3$)&
    (c) Badnets ($5\times 5$)\\
    \end{tabular}
    \vspace{-3mm}
    \caption{Trade-off curves between accuracy and attack success rate of different blind erasing methods against different triggers. The curves are drawn by adjusting the learning rate of these methods (except DHBE) from 0.002 to 0.02. } 
    \label{fig:trade_off_plot_cifar10}
\end{figure*}

\subsubsection{Configurations for Backdoor Erasing Methods:}
Since our framework is the first attempt in data-free backdoor erasing, we compare its performance with existing data-driven backdoor erasing methods. Specially, we categorize those data-driven methods into blind and targeted erasing methods, and provide detailed comparison experiments with these two kinds of methods respectively.

\noindent
\textbf{\quad (a) Blind Erasing Methods:}
We compare our DHBE framework with four existing blind erasing methods: 
1) finetuning, 2) finepruning \cite{liu2018fine}, 3) mode connectivity repair (MCR) \cite{zhao2019bridging} and 4) neural attention distillation (NAD) \cite{li2021neural}.
For these data-driven methods, we assume that they all have access to the same 4\% of clean training data (2000 samples).

\noindent
\textbf{\quad (b) Targeted Erasing Methods:}
Targeted erasing methods need to determine the attacked class first, then perform the trigger recovery and backdoor erasing process. For a fair comparison on backdoor erasing effectiveness, we assume that those methods are already known the attacked class, then trigger recovery and backdoor erasing are performed on the attacked class.
Two targeted erasing methods are employed: NC \cite{wang2019neural} and GDM \cite{qiao2019defending}.

\noindent
\textbf{\quad (c) Proposed DHBE:}
We designed two generators in DHBE: the sample generator $\mathcal{G}$ and the trigger generator $\mathcal{G}_p$ ($h=w=5,s=1.0$).
The detailed design of these two generators is described in Appendix. Only one hyperparameter is included in DHBE's loss functions : $\lambda$ in Eq. \ref{eq:adversarial_obj_function}, which controls the degree of adversarial backdoor regularization. 
We set $\lambda$ to 0.1 in all experiments. 
Further analysis of this hyperparameter is included in ablation studies.
For optimizers, we globally employ an SGD optimizer with initial learning rate of 0.1, momentum of 0.9, and weight decay of 5e-4 to update the student model, and use an Adam optimizer with initial learning rate of 1e-3 to update the generators. The student and generators are jointly optimized for 50 iterations $\times$ 300 epochs, where the student is updated by three times and generators are updated once in one iteration. 128 fake samples and triggers are generated in each iteration. The learning rates of these optimizers are simultaneously decayed by 0.1 at epoch 180 and 240. The whole backdoor erasing process takes about 6h on a GTX 2080TI gpu.

\subsubsection{Evaluation Metrics}
Two metrics are employed to evaluate the quality of backdoor erasing methods: ACC and ASR, For each backdoor erasing method, ACC and ASR are calculated using the cleaned model,  and are compared with that of the original backdoored model to judge its performance. A good backdoor erasing method should not only mitigate the ASR metric, but also keep the ACC metric the same as the original backdoored model. 

\noindent
\textbf{\quad (a) Accuracy on Clean Data (ACC)}. The test accuracy on the benign test set, which shows the influence of the backdoor erasing process on the functionality of the model. 

\noindent
\textbf{\quad (b) Attack Success Rate (ASR)}.
the ASR of model $\mathcal{S}$ is defined as the fraction of correctly classified inputs that are not labeled as the target class but misclassified to the target class after the backdoor trigger is injected:
\begin{equation}
    ASR(\mathcal{S}) = P \left(\mathcal{S}(x^\delta, \theta_s) = t \bigm| y \neq t, \mathcal{S}(x, \theta_s) = y \right),
\end{equation} 
where $(x, y)$ is a sample from the test dataset and $x^\delta$ is the input that has been injected with a trigger pattern $\delta$. $t$ is the target class and $\theta_s$ is the weights of the model $\mathcal{S}$. Note that this metric only relies on the test data that is correctly classified by the backdoored model.

\subsection{Experimental Results}
Due to the experimental settings are already complicated by various attack methods and different triggers, in this section, we focus on the CIFAR10 classification task first, and present detailed comparison results and analyses using Resnet18 models trained on the CIFAR10 dataset and attacked by different methods and triggers.
We first compare DHBE with existing blind erasing methods. 
Then, for those targeted erasing methods whose effectiveness are dependent on correctly discover the attacked class, we omit their diagnosing process, directly employ their trigger recovery routines and erasing routines on the backdoored model and report their backdoor erasing performance.

\subsubsection{Comparison with Blind Erasing Methods}
The comparison results of our framework with four blind erasing methods on different kinds of backdoor attacks are shown in Table \ref{tab:results_cifar10_small_trigger_blinded}. As shown in Table \ref{tab:results_cifar10_small_trigger_blinded}, our framework outperforms other methods by a large margin on all kinds of backdoor attacks: DHBE only sightly degrades the performance of the original model (less than $1\%$), and reduce the attack success rate of all triggers to nearly neglectable.
In contrast, the results of Finetuning, Finepruning, and NAD has about 4\% - 5\% accuracy degradation when the learning rate is set to $0.01$. Under this setting, the backdoor elimination effectiveness of Finetuning and Finepruning is unstable and failed to suppress ASR below 10\% on some triggers.
The results of NAD are relatively more stable, the ASRs of different triggers are all suppressed below 10\%, but NAD still greatly degrades the model's accuracy. 
We note that NAD is claimed to have about 2.5\% accuracy degradation in their experiments, however, they employ weak baselines where the reported accuracy is below the model's true capabilities. In our experiments, we use models that are trained with moderate data augmentations so that the backdoored model's accuracy reflects the true capabilities of the model. Therefore, the accuracy decay we report is closer to real-world scenarios.

\begin{table*}[h]
    \small
    \centering
    \caption{Results of DHBE and other methods on other datasets against Badnets attack with different triggers. Numbers are displayed as percentages. ``$\dagger$" represents the attacked class $t$ is provided for the method to perform backdoor erase on.}
    
	\begin{tabular}{c|c|cc|cc|cc|cc|cc|cc}
	\toprule
	\multirow{3}{*}{Datasets} & 
    \multirow{3}{*}{  \shortstack{Trigger\\Size}} &  
		\multicolumn{2}{c|}{Backdoored} & 
			\multicolumn{2}{c|}{Finetuning} &
				\multicolumn{2}{c|}{NAD \cite{li2021neural}} &
					\multicolumn{2}{c|}{NC$^\dagger$ \cite{wang2019neural}} &
						\multicolumn{2}{c|}{GDM$^\dagger$ \cite{qiao2019defending}} &
							\multicolumn{2}{c}{DHBE} \\
	&	&
	 	\multicolumn{2}{c|}{$N_{poison}=300$}	& 
			\multicolumn{2}{c|}{$N_{clean}=2000$} & 
				\multicolumn{2}{c|}{$N_{clean}=2000$} & 
					\multicolumn{2}{c|}{$N_{clean}=1000$} & 
						\multicolumn{2}{c|}{$N_{clean}=1000$} & 
							\multicolumn{2}{c}{No data required} \\
	&	& 
		ACC & ASR & 
			ACC & ASR & 
				ACC & ASR & 
					ACC & ASR & 
						ACC & ASR &
							ACC & ASR \\
	\hline
	\multirow{2}{*}{CIFAR100} & $3 \times 3$ &
 77.30 & 99.99  &  68.55 & 86.51  &  70.01 & 83.47  &  74.63 & 0.70  &  76.00 & \textbf{0.12}  &  \textbf{76.06} & 0.66  \\
	& $5 \times 5$ & 
 77.29 & 99.83  &  68.49 & 91.48  &  69.54 & 81.39  &  75.39 & 34.70  &  \textbf{76.11} & \textbf{0.04}  &  75.17 & 2.58  \\
	\hline
	\multirow{2}{*}{Mini-Imagenet} & $3 \times 3$ & 
 68.61 & 98.56  &  58.36 & 11.91  &  60.13 & 42.26  &  66.55 & 0.08  &  \textbf{68.05} & \textbf{0.06}  &  66.61 & 1.18  \\ 
	& $5 \times 5$ &
 68.51 & 99.15  &  56.37 & 18.10  &  59.09 & 45.70  &  65.00 & 2.00  &  \textbf{68.31} & \textbf{0.10}  &  64.29 & 2.60  \\
	\hline
	\multirow{2}{*}{VGGFace2} & $3 \times 3$ &
 96.85 & 99.97  &  91.49 & 4.35  &  91.91 & 3.35  &  95.91 & 0.01  &  \textbf{96.43} & \textbf{0.00}  &  96.32 & 6.49  \\			
	& $5 \times 5$ &
 96.75 & 99.97  &  92.19 & 14.74  &  92.39 & 5.03  &  95.31 & 97.05  &  \textbf{96.06} & \textbf{0.00}  &  95.84 & 0.01  \\
	\bottomrule	
	\end{tabular}

    \label{tab:results_other_ds_small_trigger}
\end{table*}

Despite the weak performance of those blind erasing methods, their effectiveness seems to be extremely sensitive to hyperparameters and the quantity of the clean dataset. 
To clearly demonstrate how sensitive those blind erasing methods are to the aforementioned two factors, we show the ACC-ASR trade-off curves of those methods in Fig. \ref{fig:trade_off_plot_cifar10}:
For each method, we run the method on the backdoored model multiple times with the experimental learning rate linearly increases from 0.002 to 0.02, and record the ACC and ASR of the resulted model for each run. We then draw the scatter plot of these results, and use a logistic model to fit the scattered points as the expected ACC-ASR trade-off curve of the evaluated method.
Specifically, we evaluate the ACC-ASR trade-off curve of those methods two times with 4\% of clean data and 1\% of clean data.
For each curve, five different random subsets of the clean dataset are used in evaluation.
From Fig. \ref{fig:trade_off_plot_cifar10}, the effectiveness and drawbacks of existing blind erasing methods are clearly demonstrated:

\begin{table}[t]
    \centering
    \caption{Comparison results of DHBE to targeted erasing methods on CIFAR10 against patch triggers. Numbers are displayed as percentages. ``$\dagger$" represents the attacked class $t$ is provided for the method to perform backdoor erase on.}
    
    \small
    
\scalebox{0.9}{
\begin{tabular}{c|c|cc|cc|cc|cc}
	\toprule
		 \multirow{3}{*}{\shortstack{Att.\\Meth.}} &
		\multirow{3}{*}{Tri.} &
			\multicolumn{2}{c|}{Backdoored} & 
				\multicolumn{2}{c|}{NC$^\dagger$ \cite{wang2019neural}} &
					\multicolumn{2}{c|}{GDM$^\dagger$ \cite{qiao2019defending}} &
						\multicolumn{2}{c}{DHBE}  \\
		&  	&  
			\multicolumn{2}{c|}{$t$=``truck"} &
				\multicolumn{2}{c|}{$N_{cl.}=1000$} &
					\multicolumn{2}{c|}{{$N_{cl.}=1000$}} &
						\multicolumn{2}{c}{No data req.}  \\
		& 	&  
			ACC & ASR &  
				ACC & ASR &  
					ACC & ASR &  
						ACC & ASR  \\
	\hline
	\multirow{3}{*}{  \shortstack{B.\\ \cite{gu2017badnets}} }
		& $2\times 2$ &  
 94.75 & 100.00  &  93.27 & 6.96  &  93.98 & 0.98  &  \textbf{94.05} & \textbf{0.02}  \\
	\cline{2-10}
		& $3\times 3$ &  
 95.02 & 100.00  &  93.47 & 17.83  &  94.23 & \textbf{0.06}  &  \textbf{94.46} & 0.12  \\
	\cline{2-10}
		& $5\times 5$ & 
 95.11 & 100.00  &  93.94 & 0.66  &  94.18 & 0.11  &  \textbf{94.24} & \textbf{0.00}  \\
	\hline
	\multirow{3}{*}{  \shortstack{C.\\ \cite{turner2019label}} }
		& $2\times 2$ &  
 94.85 & 46.32  &  93.32 & 21.13  &  \textbf{94.44} & 0.48  &  94.19 & \textbf{0.14}  \\
	\cline{2-10}
		& $3\times 3$ &  
 95.04 & 93.45  &  93.35 & 54.45  &  \textbf{94.52} & \textbf{0.37}  &  94.11 & 1.17  \\
	\cline{2-10}
		& $5\times 5$ & 
 95.25 & 100.00  &  93.40 & 73.79  &  \textbf{94.58} & \textbf{0.13}  &  94.45 & 4.48  \\
	\hline
	\multirow{3}{*}{  \shortstack{T.\\ \cite{liu2017trojaning}} }
		& $2\times 2$ & 
 94.56 & 56.20  &  93.02 & 1.44  &  93.73 & \textbf{0.14}  &  \textbf{93.96} & 4.39  \\
	\cline{2-10}
		& $3\times 3$ & 
 95.00 & 89.91  &  92.91 & 0.58  &  93.93 & \textbf{0.08}  &  \textbf{94.40} & 2.88  \\
	\cline{2-10}
		& $5\times 5$ & 
 94.93 & 99.91  &  94.20 & 0.14  &  94.09 & \textbf{0.11}  &  \textbf{94.56} & 1.13  \\

	\hline

	\multicolumn{2}{c|}{Mean} & 
 94.95 & 87.31  &  93.43 & 19.66  &  94.19 & \textbf{0.27}  &  \textbf{94.27} & 1.59  \\
	\bottomrule
	
	\end{tabular}
}

    \vspace{-6mm}
    \label{tab:results_cifar_small_trigger_target}
\end{table}

\begin{figure}[t]
    \centering
    \begin{tabular}{cc}
    \includegraphics[width=0.22\textwidth]{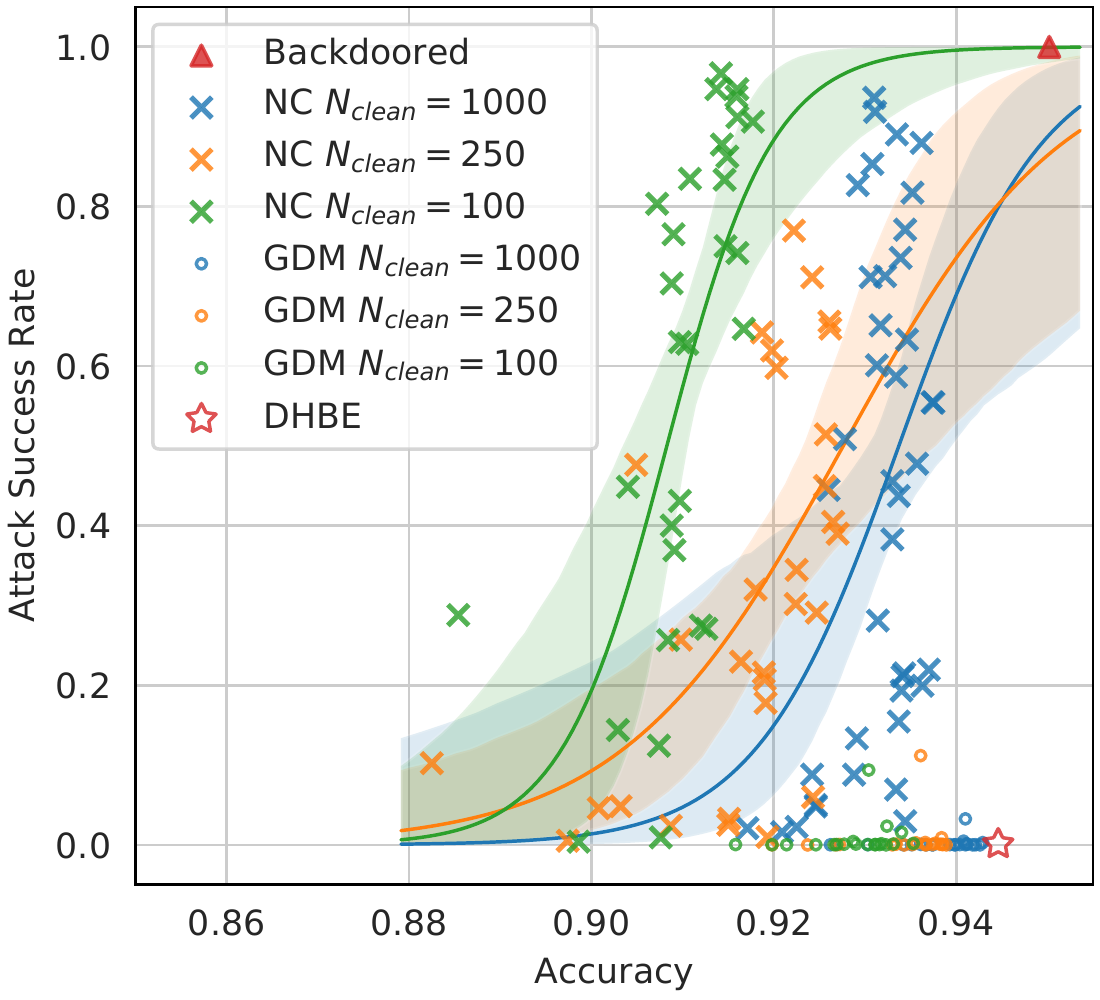} &
    \includegraphics[width=0.22\textwidth]{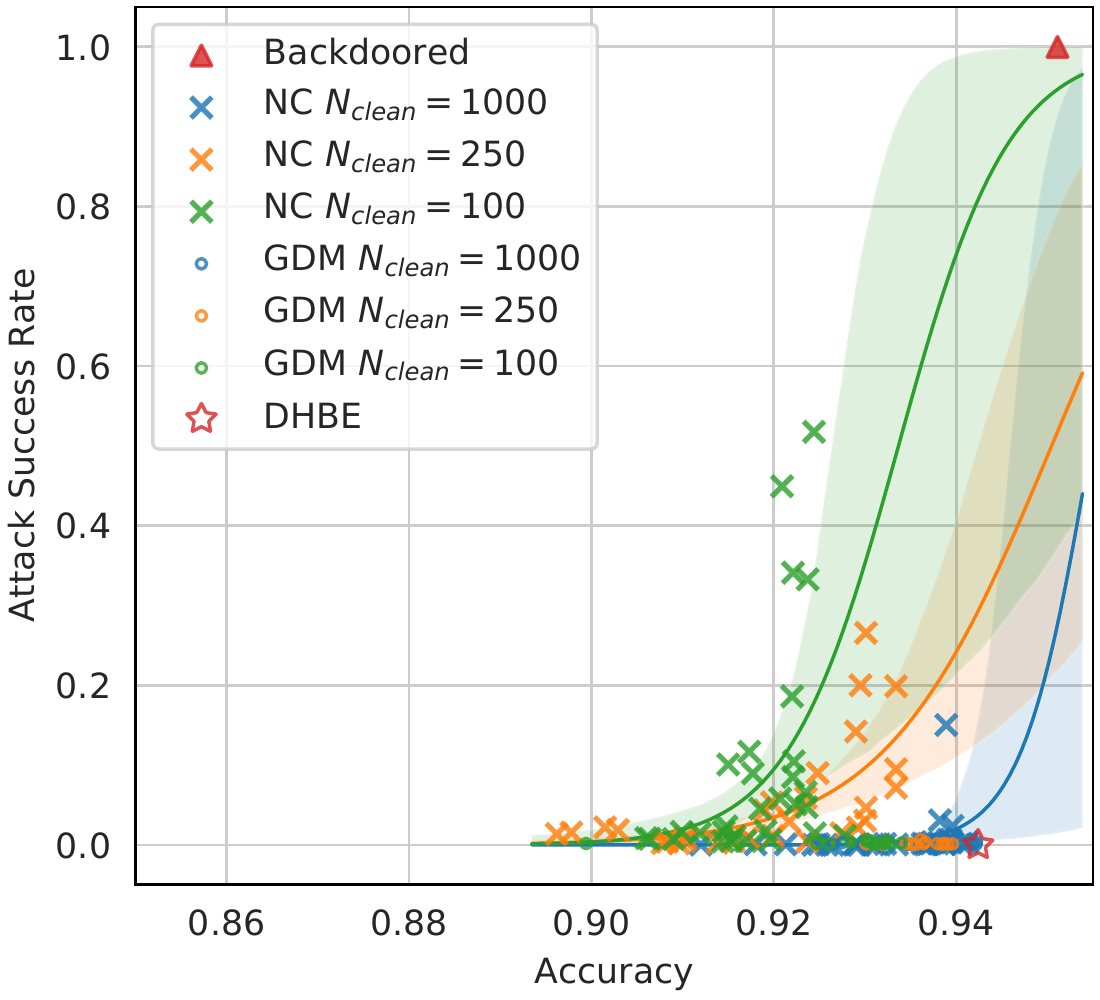}\\
    (a) Badnets ($3\times 3$) &
    (b) Badnets ($5\times 5$)
    \end{tabular}
    \vspace{-3mm}
    \caption{Trade-off curves between accuracy and attack success rate of different targeted erasing methods against different triggers. The curves are drawn by adjusting the learning rate of these methods (except DHBE) from 0.0002 to 0.002.} 
    \vspace{-5mm}
    \label{fig:trade_off_plot_cifar10_targeted}
\end{figure}

\noindent
\textbf{\quad (a) Blind erasing methods are extremely sensitive to the learning rate and the quantity of the clean dataset.} The points of finetuning, finepruning, and NAD methods in figures are drawn with learning rates in $[0.002,0.02]$. As shown in the figures, a large learning rate could mitigate ASR to nearly neglectable, but severe accuracy degradation is also observed.
However, if a small learning rate is employed, the resulted model still can be attacked with a high ASR.
The balancing of the trade-off is hard for the defender since the ASR cannot be measured in real-world applications. 
Another problem for these blind erasing methods is the quality dataset, when an adequate number of clean data is accessible ($N_{clean}=2000$), those methods provide moderate backdoor erasing effectiveness at accuracy costs of about $7\%$. But when the number of clean data is more scarce ($N_{clean}=500$), the accuracy cost is too much to afford (about $15\%$).
In contrast, the DHBE framework is insensitive to hyperparameters (learning rate, $\lambda$) due to its adversarial design, which is demonstrated in ablation studies, and DHBE framework do not require any clean data for backdoor erasing.

\noindent
\textbf{\quad (b) Blind erasing methods become less effective as the trigger size increases.}
When the trigger size increases from $2\times 2$ to  $5 \times 5$, the blind erasing methods achieve less model accuracy when the ASR is suppressed to the same level. This phenomenon is clearly demonstrated as we show the ACC-ASR curves of $2\times2$, $3\times 3$ , $5 \times 5$ triggers in Figure \ref{fig:trade_off_plot_cifar10} (a), (b), (c), respectively.
For triggers of size $2\times 2$, unlearning methods lost about 5\% model accuracy, but for triggers of size $5\times 5$, they lost 8\% model accuracy.
This may suggest that more neurons are influenced by triggers of large size, causing it hard to be erased by the unlearning methods. 
In contrast, the proposed DHBE framework appears to be equally effective to different size of triggers. 

\begin{table*}[htb!]
    \centering
    \caption{Comparison results of DHBE to other backdoor erasing methods against amplified backdoor attacks. Numbers are displayed as percentages. }
    \scalebox{0.91}{
	\begin{tabular}{c|ccc|ccc|ccc|ccc|ccc|ccc}
	\toprule
		\multirow{2}{*}{  \shortstack{Trigger\\Size}  } &  
			\multicolumn{3}{c|}{Finetuning} &
				\multicolumn{3}{c|}{Finepruning \cite{liu2018fine}} &
					\multicolumn{3}{c|}{NAD \cite{li2021neural}} &
						\multicolumn{3}{c|}{NC \cite{wang2019neural}} &
							\multicolumn{3}{c|}{GDM \cite{qiao2019defending}} &
								\multicolumn{3}{c}{DHBE}  \\

	\cline{2-19}
		 & 
			 x1 & x4 & x9 & 
				 x1 & x4 & x9 & 
					 x1 & x4 & x9 & 
						 x1 & x4 & x9 & 
							x1 & x4 & x9 & 
								x1 & x4 & x9 \\

	\hline
		$2\times 2$ &
 2.64 & 50.68 & 93.08  &  12.92 & 92.88 & 99.62  &  0.38 & 7.30 & 12.92  &  6.96 & 44.04 & 54.85  &  0.98 & 9.15 & 16.86  &  0.02 & 0.51 & 1.94  \\

		$3\times 3$ &
 2.97 & 37.98 & 66.32  &  6.74 & 63.47 & 98.34  &  7.08 & 84.34 & 100.00  &  17.83 & 52.95 & 83.43  &  0.06 & 0.18 & 1.89  &  0.12 & 0.22 & 1.00  \\

		$5\times 5$ &
 1.10 & 4.75 & 22.11  &  69.07 & 100.00 & 100.00  &  3.26 & 28.70 & 73.81  &  1.20 & 21.33 & 78.08  &  0.18 & 0.63 & 16.15  &  0.00 & 0.09 & 0.66  \\

	\cline{1-19}
		Mean ASR &
 2.24 & 31.14 & 60.50  &  29.58 & 85.45 & 99.32  &  3.57 & 40.11 & 62.24  &  8.66 & 39.44 & 72.12  &  0.41 & 3.32 & 11.63  &  
    \textbf{0.05} & \textbf{0.28} & \textbf{1.20}  \\

	\bottomrule
	\end{tabular}
}

    \label{tab:table_amplified_trigger}
\end{table*}

\begin{figure*}[t]
    \centering
    \vspace{-2mm}
    \includegraphics[width=0.90\textwidth]{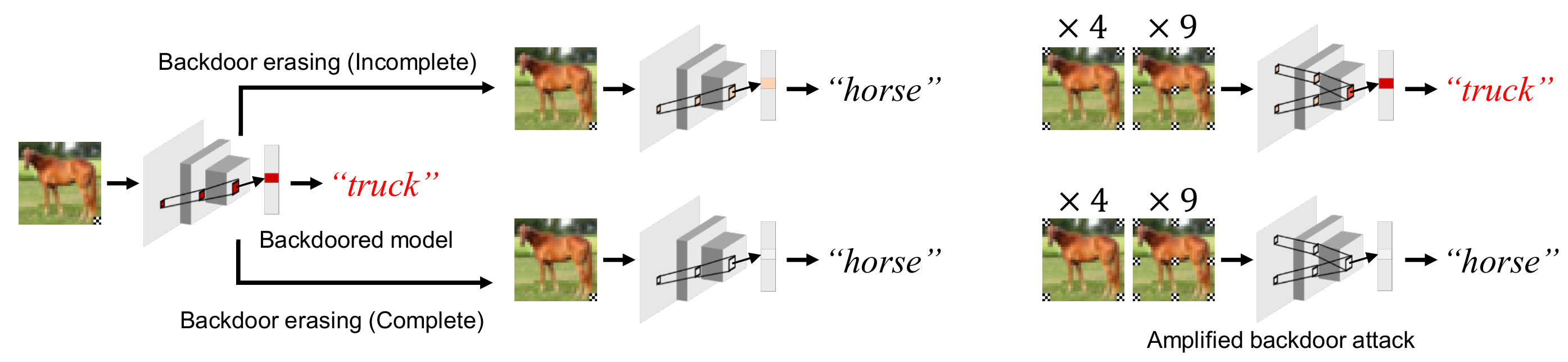}
    \vspace{-3mm}
    \caption{Illustration of how amplified backdoor attacks could still backdoor the models produced by backdoor erasing methods where injected backdoors are incompletely erased.} 
    \vspace{-3mm}
    \label{fig:amplified_backdoor_attacks}
\end{figure*}

\begin{figure}[t]
    \centering
    \includegraphics[width=0.49\textwidth]{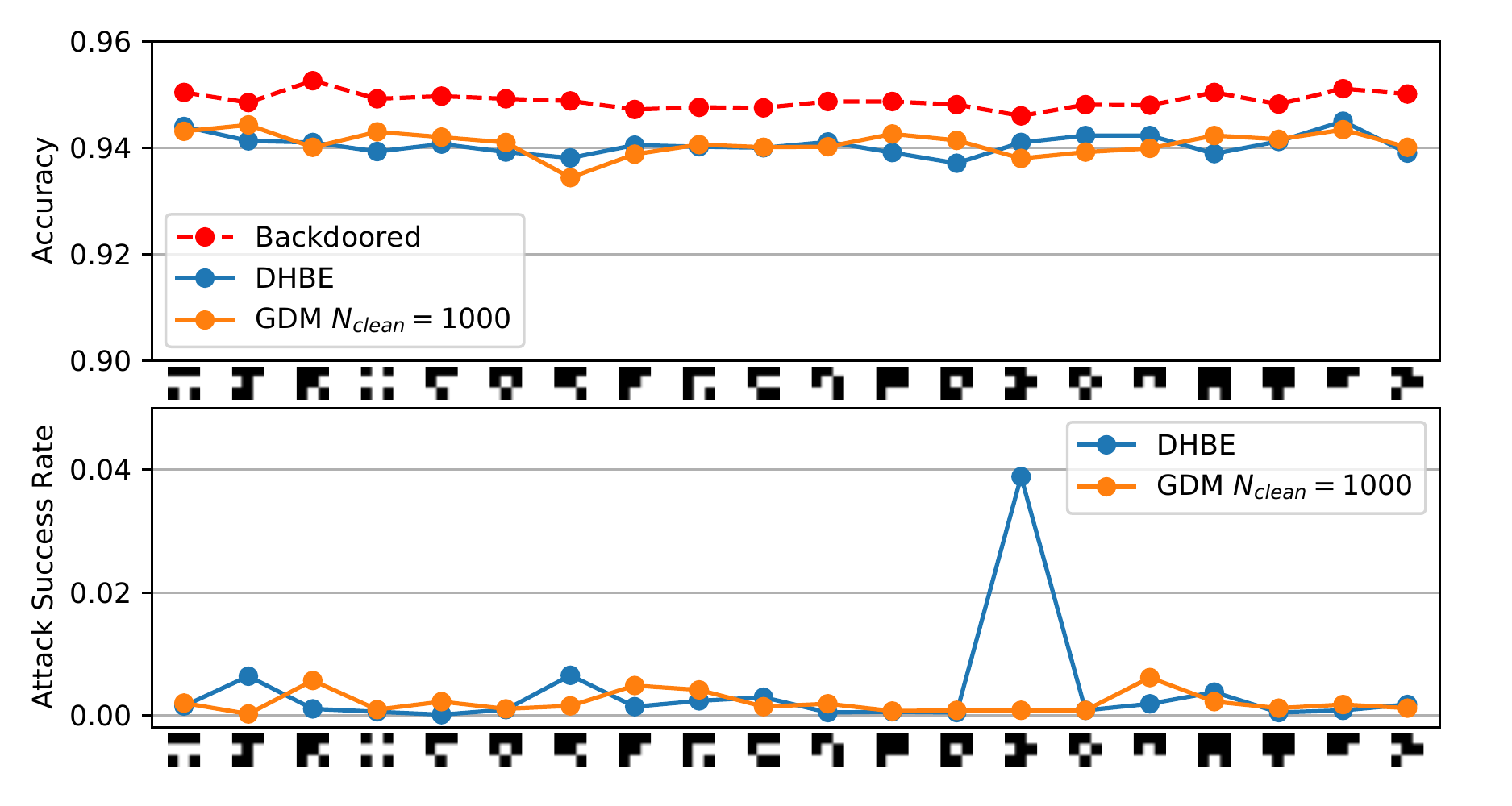}
    \vspace{-6mm}
    \caption{Comparison experiments of ABR and GDM on various triggers. ASRs of backdoored models are all close to 100\% and omitted in the bottom figure.}
    \vspace{-4mm}
    \label{fig:abr_vs_gdm}
\end{figure}

\subsubsection{Comparison with Targeted Erasing Methods} 
We also evaluate NC \cite{wang2019neural} and GDM \cite{qiao2019defending} in the same experimental settings. The results are shown in Table \ref{tab:results_cifar_small_trigger_target}: DHBE achieves comparable results with Neural Cleanse and GDM against all backdoor attacks. 
We also show ACC-ASR trade-off curves of those targeted erasing methods in Fig. \ref{fig:trade_off_plot_cifar10_targeted}.
As shown in those figures, NC and GDM are still sensitive to the employed learning rate, but they achieve much better results than blind erasing methods.

\textbf{Comparison with Generative Distribution Modeling:}
Qiao \textit{et al.} \cite{qiao2019defending} discovers that a backdoored model can be triggered by a distribution of triggers, not only a single trigger. Thus, erasing a single recovered trigger using Neural Cleanse is not robust enough. In GDM \cite{qiao2019defending}, a sampling-free distribution modeling for valid triggers was proposed, then the backdoor erasing is performed by erasing all triggers within this distribution. GDM demonstrates its consistent effectiveness against all $3\times 3$ triggers.
In the DHBE framework, owing to the adversarial optimization, the trigger generator $\mathcal{G}_p$ only needs to recover the currently most sensitive trigger to the student model $\mathcal{S}$. As $\mathcal{S}$ and $\mathcal{G}_p$ are updated adversarially and simultaneously, any trigger that can be generated by $\mathcal{G}_p$ will be mitigated.
In Fig. \ref{fig:abr_vs_gdm}, we perform DHBE and GDM \cite{qiao2019defending} on a large number of $3\times 3$ square triggers, and show that the adversarial backdoor regularization has comparable performance with GDM, even it does not model the trigger distribution.

\subsubsection{Evaluations on More Datasets}

To show that the proposed framework could be easily extended to datasets with larger size and more categories, we run the DHBE on three other datasets using the same hyperparameters and learning rates : CIFAR100, Mini-Imagenet, and VGGFace2, and list the results in Table \ref{tab:results_other_ds_small_trigger}. Specifically, for VGGFace2, we randomly choose 100 identities and train a backdoored classification model on them. We also list the results of finetuning and NAD with learning rate of 0.001 and $N_{clean}=2000$ in the table. 
The results show that our method still outperforms the blind erasing methods by a large margin.

\subsection{Incomplete Backdoor Erasing is Vulnerable under Amplified Backdoor Attack}

Existing backdoor erasing methods default that the defense is sufficient if the attack success rate is suppressed below a given threshold, such as 10\% \cite{li2021neural}. However, in this work, we demonstrate that a simple attack strategy could easily trigger the cleaned model and obtain a high attack success rate, as shown in Fig. \ref{fig:amplified_backdoor_attacks}.
The attack strategy is apparent and straight-forward:
If the injected triggers are not wiped out completely, the cleaned models still have obvious responses to the triggers, then the attacker can inject the same trigger multiple times into the input to amplify the activation of backdoors. 
The activations of multiple triggers are accumulated in the model, finally, the model's prediction can still be altered by the attacker with a high attack success rate.
Thus, it can be concluded that incomplete backdoor erasing is actually useless in real-world applications since it cannot defend against the intentional manipulation of unknown adversaries.

Here, we conduct experiments where test inputs are injected with triggers 4 or 9 times, and the attack samples are shown in Fig. \ref{fig:amplified_backdoor_attacks}. To amplify the activations of backdoors in the cleaned model, We first inject triggers into four corners of the test inputs ($\times 4$), then we further inject more triggers into the center and four edges of the test inputs ($\times 9$). 
The ASR results are shown in Table \ref{tab:table_amplified_trigger}. 
By injecting four triggers into the four corners of the test inputs, the attack remains highly stealthy.
However, the cleaned models produced by Finetuning, Finepruning, and NAD show a high attack success rate in most cases, even though the model's ASR on a single trigger is suppressed below 10\%. When more triggers are injected, those cleaned models could be attacked with nearly 100\% ASR in many cases.
Thus, we conclude that those unlearning methods are unsafe if the injected backdoors are not completely cleaned.
Since the attacker still can activate the backdoors with a high ASR, their backdoor erasing attempt is useless.
In contrast, the DHBE framework mitigates the response of triggers in the resulted model completely using the proposed adversarial backdoor regularization, and demonstrates its robustness against amplified backdoor attacks.

\begin{figure}[t]
    \centering
    \includegraphics[width=0.48\textwidth]{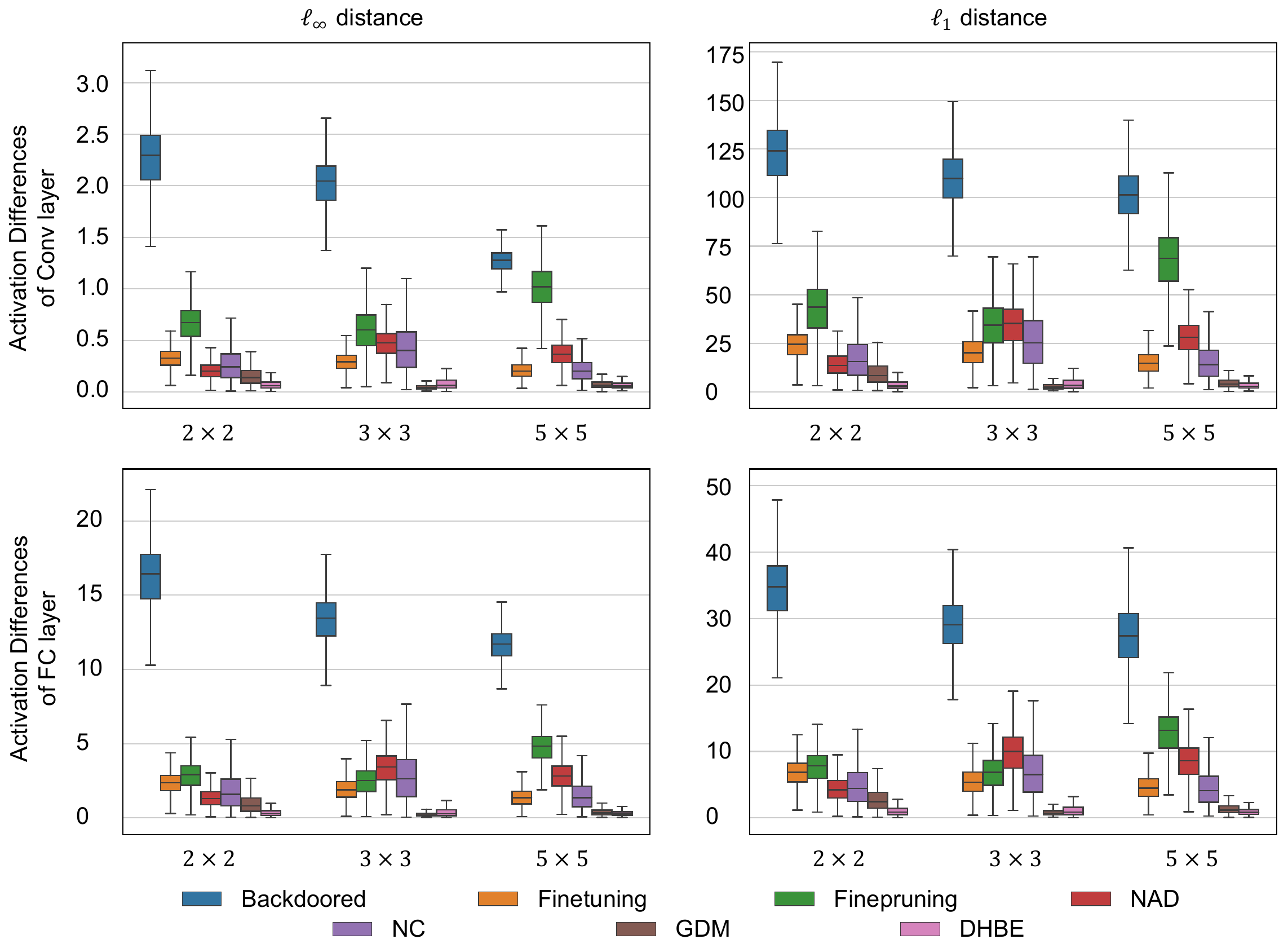}
    \vspace{-4mm}
    \caption{Activation differences between clean inputs and trigger-pasted inputs on different models. The activation differences of the last $conv$ and $fc$ layers are shown above and below, respectively. The left figures show $\ell_\infty$ distances and the right figures show $\ell_1$ distances. The models produced by DHBE always has the minimal response to the triggers.}
    \vspace{-7mm}
    \label{fig:activation_difference}
\end{figure}

To quantitatively measure the effectiveness of mitigating the model's backdoor reactions, we draw samples from the clean test dataset, and plot the distribution of activation differences between clean inputs and corresponding trigger-pasted inputs. The results are shown by boxplot in Fig. \ref{fig:activation_difference}. The activation differences of the last $conv$ layer and the final $fc$ output layer are included, and differences are calculated by two measures : $\ell_1$ distance and $\ell_\infty$ distance. As shown in Fig. \ref{fig:activation_difference}, the model produced by DHBE demonstrates far less response to the triggers than other methods
even when measured at an inner layer.

\begin{figure}[t]
    \centering
    \includegraphics[width=0.48\textwidth]{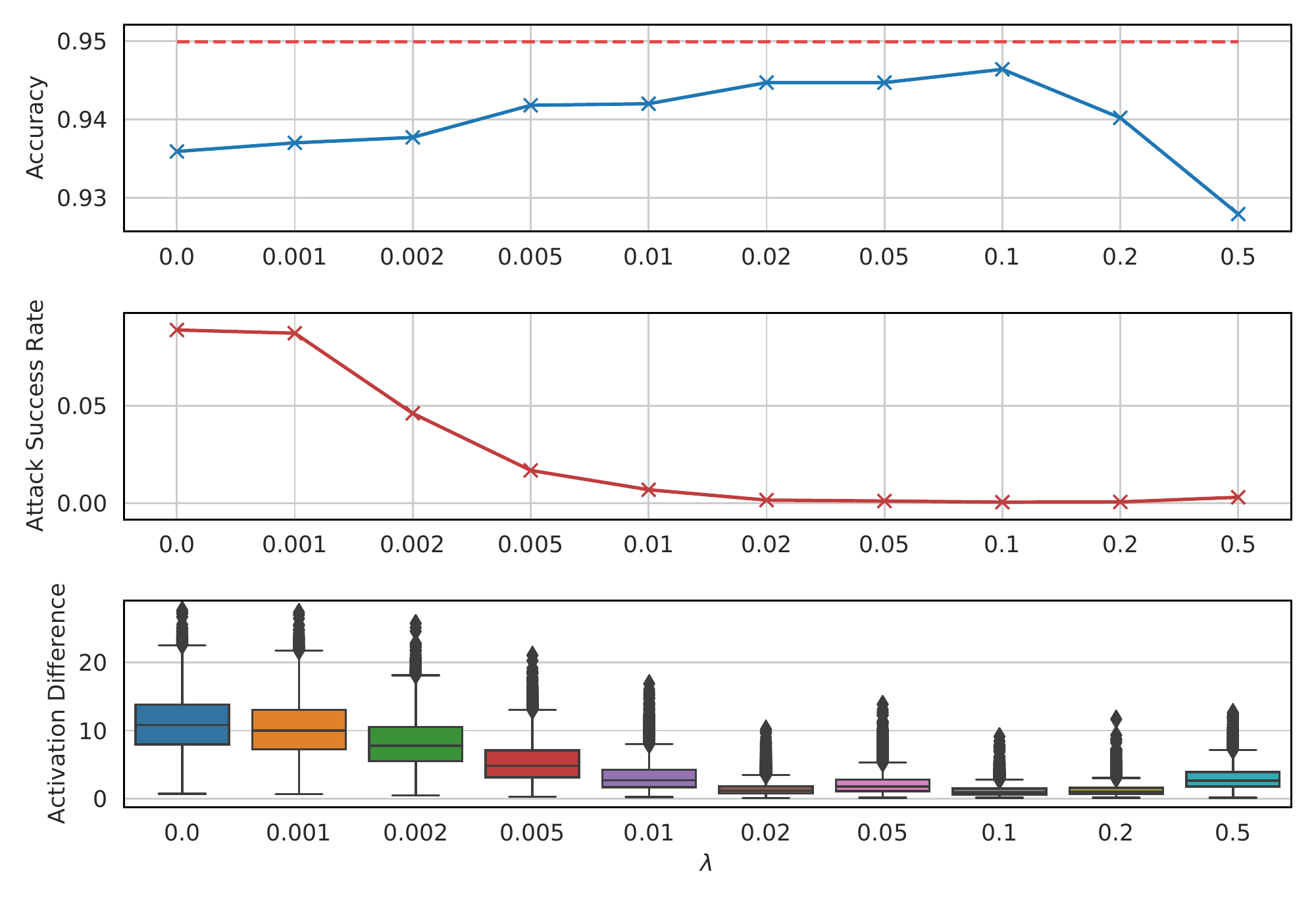}
    \vspace{-7mm}
    \caption{Ablation studies on ABR (controlled by $\lambda$) of DHBE framework (against Badnets attack with $3\times 3$ square trigger). We show the changes of accuracy, attack success rate, and activation difference (Outputs of $fc$ layer, $\ell_1$ distance) w.r.t. different $\lambda$. Red dash line in top figure represents the accuracy of orignal backdoored model. }
    \vspace{-3mm}
    \label{fig:ablation_abr}
\end{figure}

\begin{figure}[t]
    \centering
    \includegraphics[width=0.49\textwidth]{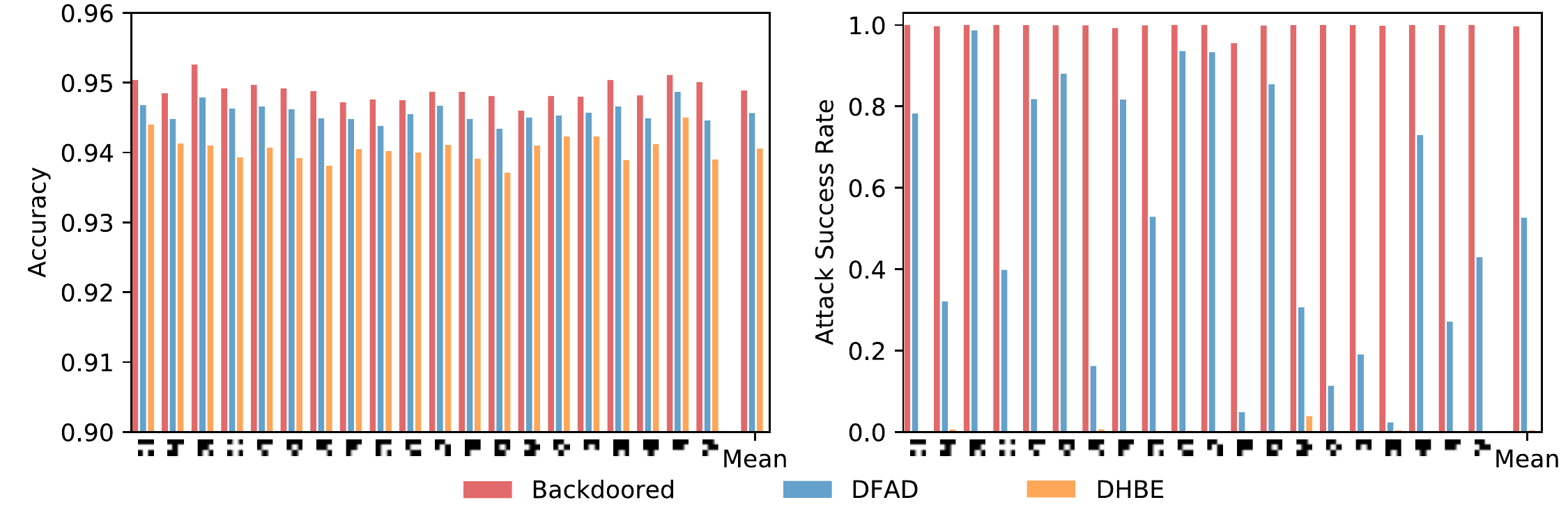}
    \vspace{-5mm}
    \caption{Comparison experiments of DHBE and pure adversarial distillation on different backdoor triggers.}
    \vspace{-5mm}
    \label{fig:dfad_vs_dhbe}
\end{figure}

\subsection{Ablation Studies}

In this subsection, we show that the effectiveness of the proposed DHBE is insensitive to a wide range of choices of hyperparameters, and DHBE is able to deal with backdoor attacks with different size of triggers using a same set of hyperparameters. These ablation studies suggest that our backdoor erasing framework is robust enough and can be deployed in real-world applications with little trouble.

\subsubsection{Effectiveness of Adversarial Backdoor Regularization}

First, we investigate the effectiveness of proposed adversarial backdoor regularization, whose magnitude is controlled by a single hyperparameter $\lambda$.
Here, we perform experiments with different values of $\lambda$, and show the accuracy, attack success rate, and activation difference ($\ell_1$ distance of the outputs of $fc$ layer) in Fig. \ref{fig:ablation_abr}. 
As the value of $\lambda$ increases, the ASR is decreased as well as the response to the trigger, until it becomes barely noticeable.
It can be also found that this regularization term effectively suppresses the backdoor reactions and maintains the accuracy of the model across a wide range of $\lambda$ (from 0.01 to 0.2).

We note that the DHBE framework mitigates the ASR from $100\%$ to $8.9\%$ when $\lambda=0.0$. This result does not weaken the actual effectiveness of ABR because 1) DHBE with $\lambda=0.0$ does not equal pure adversarial distillation. Even if $\lambda$ is set to 0.0 in DHBE, the generated trigger-pasted inputs are fed into the student model in the training mode, which influences its batchnorm layers' statistics and resulted in suppressed response to the triggers.
2) We also evaluate pure adversarial distillation with a large amount of triggers, and show the results in Fig. \ref{fig:dfad_vs_dhbe}. It can be seen that different triggers have stochastic transferability. The evaluated triggers have up to 98\% ASR and 50\% average ASR after using pure adversarial distillation, but have nearly 0\% ASR after using the proposed DHBE method.

\subsubsection{Robustness w.r.t. Different Learning Rates}

\begin{figure}[t]
    \centering
    \includegraphics[width=0.48\textwidth]{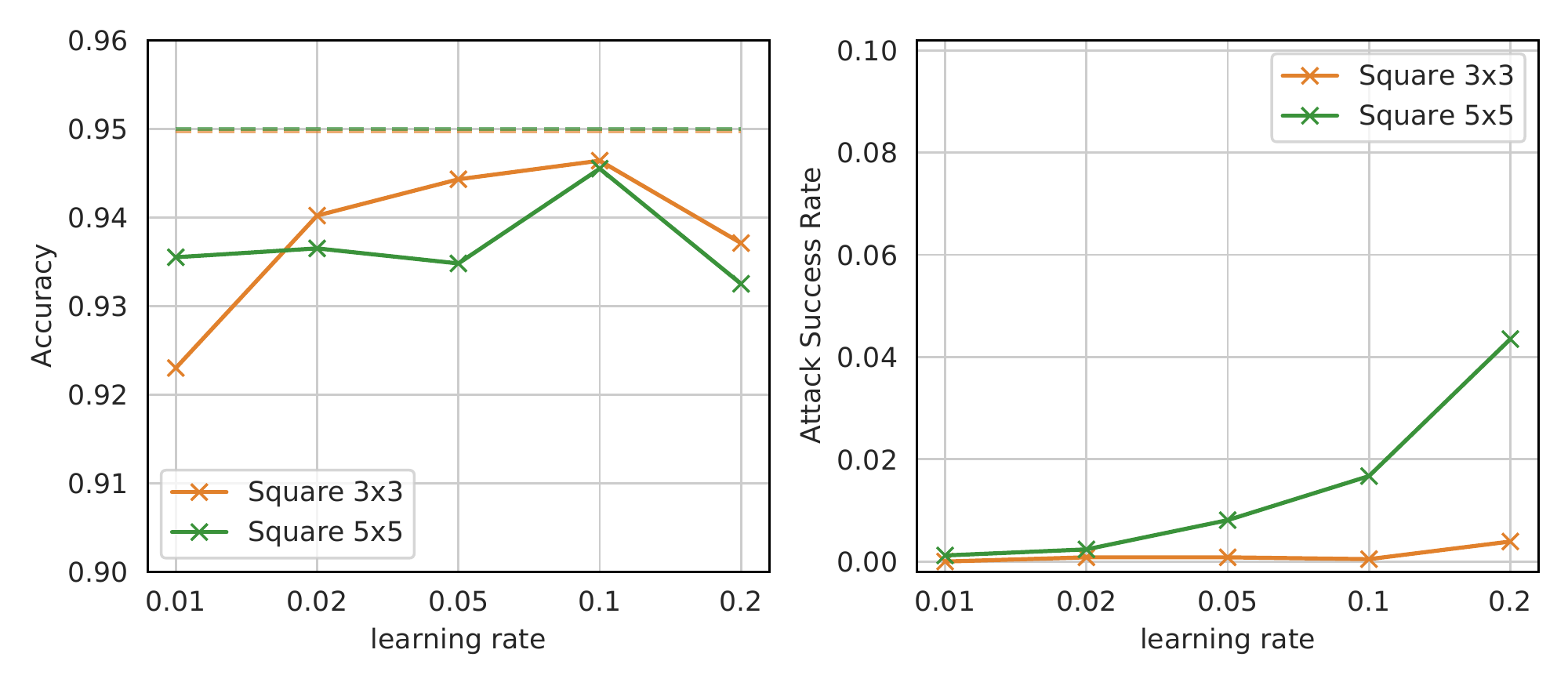}
    \vspace{-7mm}
    \caption{Ablation studies on the learning rate of DHBE framework. Dash lines in the left figure represent the accuracy of original backdoored model. ASRs of backdoored models are all close to 100\% and omitted in the right figure.}
    \vspace{-5mm}
    \label{fig:figure_ablation_learning_rate}
\end{figure}

As shown in previous experiments in Fig. \ref{fig:trade_off_plot_cifar10} and Fig. \ref{fig:trade_off_plot_cifar10_targeted}, both blind and targeted erasing methods are sensitive to the learning rate used in finetuning.
In contrast, DHBE does not need to find a proper learning rate for each backdoor erasing attempt, thanks to its adversarial design. 
The results are shown in Fig. \ref{fig:figure_ablation_learning_rate} where different learning rates from a wide range (from 0.01 to 0.2) are employed. 
The learning rates of the sample generator and trigger generator are synchronously adjusted for stabilized adversarial training. As shown in Fig. \ref{fig:figure_ablation_learning_rate}, the results of DHBE are consistent across a wide range of learning rates.

\subsubsection{Trigger size v.s. Trigger Generator Size}

\begin{figure}[t]
    \centering
    \includegraphics[width=0.48\textwidth]{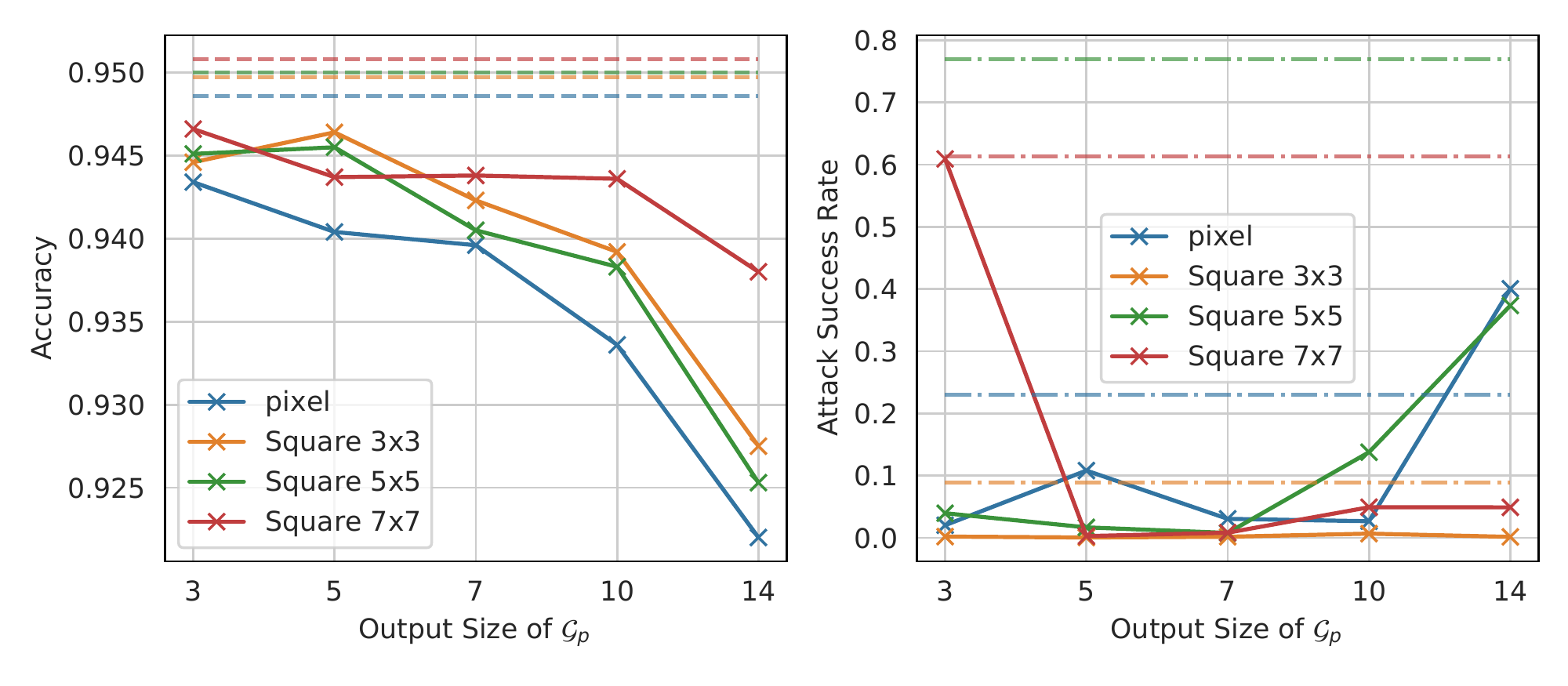}
    \vspace{-7mm}
    \caption{Ablation study on the output size of trigger generator $\mathcal{G}_p$ in DHBE framework against different size of triggers. Dash lines in the left figure represent the accuracy of original backdoored model, and dash lines in right figure represent ASRs of DHBE ($\lambda=0$) (ASRs of backdoored models are all close to 100\% and omitted in the right figure).}
    \vspace{-3mm}
    \label{fig:figure_ablation_tt_size}
\end{figure}

Although various methods have been proposed to diagnose whether a model is backdoored or not, it is still difficult for defenders to determine the size, shape, and texture of the actual trigger. In Fig. \ref{fig:figure_ablation_tt_size}, we show that the proposed DHBE framework still demonstrates effectiveness when the actual trigger size and the generation size of triggers generated by $\mathcal{G}_p$ do not well matched.
Specifically, we train backdoored models implanted with different sizes of trigger, and employ DHBE with different output sizes of $\mathcal{G}_p$ to distill them. The results in Fig. \ref{fig:figure_ablation_tt_size} show that the DHBE with trigger generator size $7\times7$ is able to erase all those triggers with only about $1\%$ accuracy degradation.

\begin{figure}[htb!]
    \centering
    \includegraphics[width=0.45\textwidth]{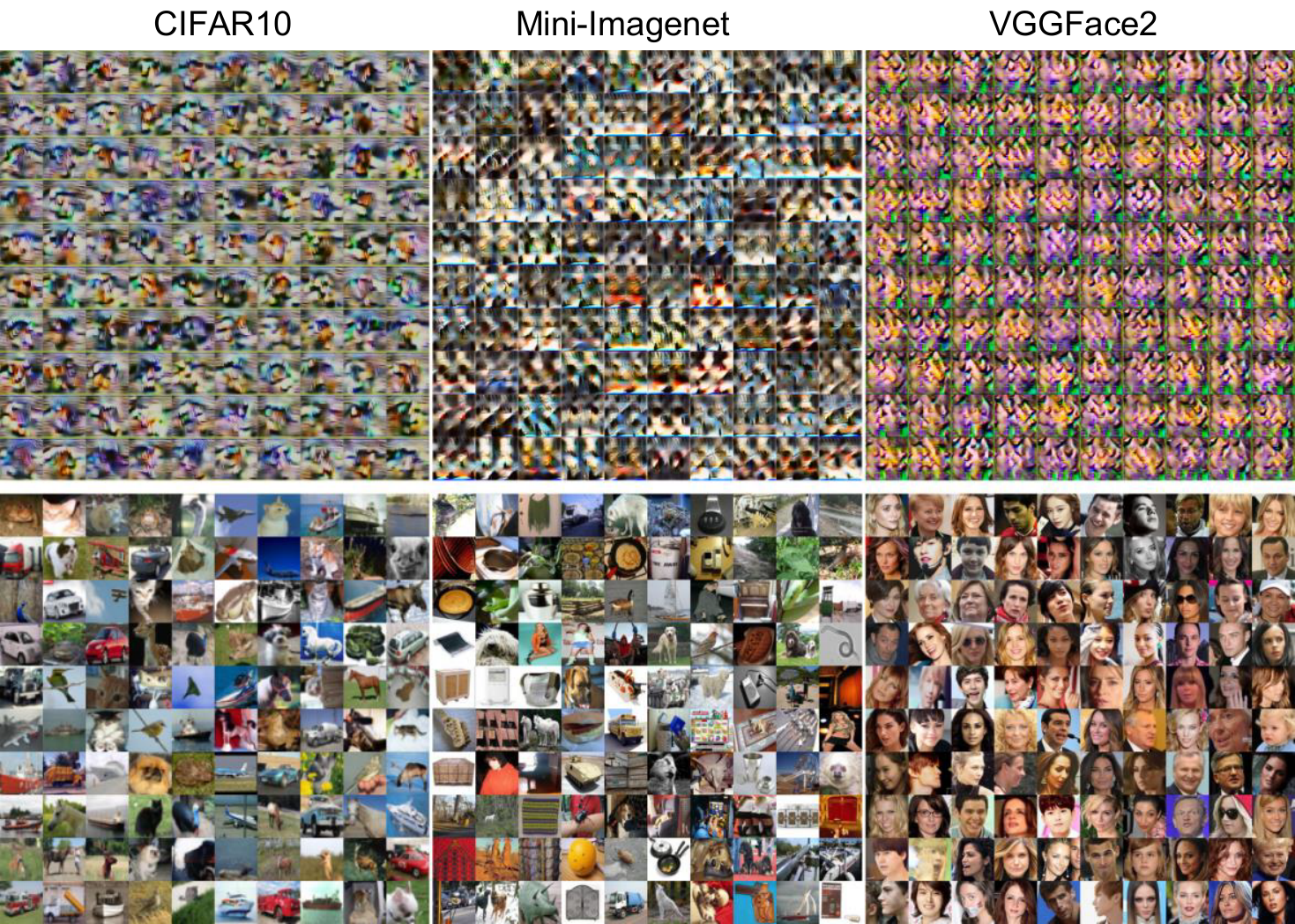}
    \caption{The first row are samples generated by $\mathcal{G}$. The images in the second row are sampled from original training data.}
    \vspace{-5mm}
    \label{fig:generated_samples}
\end{figure}

\subsubsection{Compare to Model Inverting Techniques}

A competitor for data-free adversarial distillation is distillation with inverted training dataset \cite{yin2020dreaming, yin2021see}. 
However, it will cause privacy concerns in real-world applications, as the information of the original training data is inverted and visualized.
We show that the generated samples in DHBE do not exhibit features of the original training dataset in Fig. \ref{fig:generated_samples}. This enables the DHBE to be deployed in privacy-concerning scenarios involving federated learning and so on.

Besides, the dataset inverting techniques cannot be directly combined with blind or targeted erasing methods for data-free backdoor erasing.
Our evaluations on distillation using inverted dataset \cite{yin2020dreaming, yin2021see} show similar results with pure adversarial distillation \cite{fang2019data} whose results are shown in Fig. \ref{fig:dfad_vs_dhbe}, where backdoors could be distilled and transferred into student models with a high probability. 

\section{Conclusions}

In this work, we propose a novel data-free holistic backdoor erasing framework (DHBE) to wipe out hidden triggers in attacked DNNs without access to clean data using the proposed restricted adversarial distillation paradigm.
We empirically demonstrate that the proposed framework achieves superior performance in comparison to other backdoor erasing methods, even these methods are assisted by a number of clean data.
We also quantify the models’ internal response to show that the proposed method most effectively suppresses the model's backdoor reactions.
On top of that, we explore the robustness of the proposed method, and show that it provides unprecedented stability in the tuning of hyperparameters, due to its adversarial design.
Overall, the proposed framework provides a practical, flexible, and effective solution for eliminating patch backdoors yet. 
We hope our work could inspire more comprehensive backdoor erasing efforts against evolving backdoor attacks.

\begin{acks}
This research work is funded by the National Nature Science Foundation of China under Grant 61971283, 62202303, and Shanghai Municipal Science and Technology Major Project under Grant 2021SHZDZX0102.
\end{acks}

\bibliographystyle{ACM-Reference-Format}
\bibliography{ccs-template}


\begin{thebibliography}{55}


\ifx \showCODEN    \undefined \def \showCODEN     #1{\unskip}     \fi
\ifx \showDOI      \undefined \def \showDOI       #1{#1}\fi
\ifx \showISBNx    \undefined \def \showISBNx     #1{\unskip}     \fi
\ifx \showISBNxiii \undefined \def \showISBNxiii  #1{\unskip}     \fi
\ifx \showISSN     \undefined \def \showISSN      #1{\unskip}     \fi
\ifx \showLCCN     \undefined \def \showLCCN      #1{\unskip}     \fi
\ifx \shownote     \undefined \def \shownote      #1{#1}          \fi
\ifx \showarticletitle \undefined \def \showarticletitle #1{#1}   \fi
\ifx \showURL      \undefined \def \showURL       {\relax}        \fi
\providecommand\bibfield[2]{#2}
\providecommand\bibinfo[2]{#2}
\providecommand\natexlab[1]{#1}
\providecommand\showeprint[2][]{arXiv:#2}

\bibitem[Arjovsky et~al\mbox{.}(2017)]%
        {arjovsky2017wasserstein}
\bibfield{author}{\bibinfo{person}{Martin Arjovsky}, \bibinfo{person}{Soumith
  Chintala}, {and} \bibinfo{person}{L{\'e}on Bottou}.}
  \bibinfo{year}{2017}\natexlab{}.
\newblock \showarticletitle{Wasserstein Generative Adversarial Networks}. In
  \bibinfo{booktitle}{\emph{ICML}}. \bibinfo{pages}{214--223}.
\newblock


\bibitem[Bagdasaryan and Shmatikov(2020)]%
        {bagdasaryan2020blind}
\bibfield{author}{\bibinfo{person}{Eugene Bagdasaryan} {and}
  \bibinfo{person}{Vitaly Shmatikov}.} \bibinfo{year}{2020}\natexlab{}.
\newblock \showarticletitle{Blind Backdoors in Deep Learning Models}.
\newblock \bibinfo{journal}{\emph{arXiv preprint arXiv:2005.03823}}
  (\bibinfo{year}{2020}).
\newblock


\bibitem[Barni et~al\mbox{.}(2019)]%
        {barni2019new}
\bibfield{author}{\bibinfo{person}{Mauro Barni}, \bibinfo{person}{Kassem
  Kallas}, {and} \bibinfo{person}{Benedetta Tondi}.}
  \bibinfo{year}{2019}\natexlab{}.
\newblock \showarticletitle{A New Backdoor Attack in CNNs by Training Set
  Corruption without Label Poisoning}. In \bibinfo{booktitle}{\emph{ICIP}}.
  \bibinfo{pages}{101--105}.
\newblock


\bibitem[Cao et~al\mbox{.}(2018)]%
        {cao2018vggface2}
\bibfield{author}{\bibinfo{person}{Qiong Cao}, \bibinfo{person}{Li Shen},
  \bibinfo{person}{Weidi Xie}, \bibinfo{person}{Omkar~M Parkhi}, {and}
  \bibinfo{person}{Andrew Zisserman}.} \bibinfo{year}{2018}\natexlab{}.
\newblock \showarticletitle{Vggface2: A Dataset for Recognising Faces across
  Pose and Age}. In \bibinfo{booktitle}{\emph{FG}}.
\newblock


\bibitem[Chen et~al\mbox{.}(2019a)]%
        {chen2019detecting}
\bibfield{author}{\bibinfo{person}{Bryant Chen}, \bibinfo{person}{Wilka
  Carvalho}, \bibinfo{person}{Nathalie Baracaldo}, \bibinfo{person}{Heiko
  Ludwig}, \bibinfo{person}{Benjamin Edwards}, \bibinfo{person}{Taesung Lee},
  \bibinfo{person}{Ian Molloy}, {and} \bibinfo{person}{Biplav Srivastava}.}
  \bibinfo{year}{2019}\natexlab{a}.
\newblock \showarticletitle{Detecting Backdoor Attacks on Deep Neural Networks
  by Activation Clustering}. In \bibinfo{booktitle}{\emph{AAAI}}.
\newblock


\bibitem[Chen et~al\mbox{.}(2019b)]%
        {chen2019deepinspect}
\bibfield{author}{\bibinfo{person}{Huili Chen}, \bibinfo{person}{Cheng Fu},
  \bibinfo{person}{Jishen Zhao}, {and} \bibinfo{person}{Farinaz Koushanfar}.}
  \bibinfo{year}{2019}\natexlab{b}.
\newblock \showarticletitle{DeepInspect: A Black-box Trojan Detection and
  Mitigation Framework for Deep Neural Networks.}. In
  \bibinfo{booktitle}{\emph{IJCAI}}.
\newblock


\bibitem[Chen et~al\mbox{.}(2019c)]%
        {chen2019data}
\bibfield{author}{\bibinfo{person}{Hanting Chen}, \bibinfo{person}{Yunhe Wang},
  \bibinfo{person}{Chang Xu}, \bibinfo{person}{Zhaohui Yang},
  \bibinfo{person}{Chuanjian Liu}, \bibinfo{person}{Boxin Shi},
  \bibinfo{person}{Chunjing Xu}, \bibinfo{person}{Chao Xu}, {and}
  \bibinfo{person}{Qi Tian}.} \bibinfo{year}{2019}\natexlab{c}.
\newblock \showarticletitle{Data-free Learning of Student Networks}. In
  \bibinfo{booktitle}{\emph{ICCV}}. \bibinfo{pages}{3514--3522}.
\newblock


\bibitem[Dhillon et~al\mbox{.}(2019)]%
        {dhillon2019baseline}
\bibfield{author}{\bibinfo{person}{Guneet~Singh Dhillon},
  \bibinfo{person}{Pratik Chaudhari}, \bibinfo{person}{Avinash Ravichandran},
  {and} \bibinfo{person}{Stefano Soatto}.} \bibinfo{year}{2019}\natexlab{}.
\newblock \showarticletitle{A Baseline for Few-Shot Image Classification}. In
  \bibinfo{booktitle}{\emph{ICLR}}.
\newblock


\bibitem[Doan et~al\mbox{.}(2020)]%
        {doan2020februus}
\bibfield{author}{\bibinfo{person}{Bao~Gia Doan}, \bibinfo{person}{Ehsan
  Abbasnejad}, {and} \bibinfo{person}{Damith~C Ranasinghe}.}
  \bibinfo{year}{2020}\natexlab{}.
\newblock \showarticletitle{Februus: Input Purification Defense against Trojan
  Attacks on Deep Neural Network Systems}. In
  \bibinfo{booktitle}{\emph{ACSAC}}. \bibinfo{pages}{897--912}.
\newblock


\bibitem[Fang et~al\mbox{.}(2022)]%
        {fang2022up}
\bibfield{author}{\bibinfo{person}{Gongfan Fang}, \bibinfo{person}{Kanya Mo},
  \bibinfo{person}{Xinchao Wang}, \bibinfo{person}{Jie Song},
  \bibinfo{person}{Shitao Bei}, \bibinfo{person}{Haofei Zhang}, {and}
  \bibinfo{person}{Mingli Song}.} \bibinfo{year}{2022}\natexlab{}.
\newblock \showarticletitle{Up to 100x Faster Data-Free Knowledge
  Distillation}. In \bibinfo{booktitle}{\emph{AAAI}}.
  \bibinfo{pages}{6597--6604}.
\newblock


\bibitem[Fang et~al\mbox{.}(2019)]%
        {fang2019data}
\bibfield{author}{\bibinfo{person}{Gongfan Fang}, \bibinfo{person}{Jie Song},
  \bibinfo{person}{Chengchao Shen}, \bibinfo{person}{Xinchao Wang},
  \bibinfo{person}{Da Chen}, {and} \bibinfo{person}{Mingli Song}.}
  \bibinfo{year}{2019}\natexlab{}.
\newblock \showarticletitle{Data-free Adversarial Distillation}.
\newblock \bibinfo{journal}{\emph{arXiv preprint arXiv:1912.11006}}
  (\bibinfo{year}{2019}).
\newblock


\bibitem[Goodfellow et~al\mbox{.}(2014)]%
        {goodfellow2014generative}
\bibfield{author}{\bibinfo{person}{Ian Goodfellow}, \bibinfo{person}{Jean
  Pouget-Abadie}, \bibinfo{person}{Mehdi Mirza}, \bibinfo{person}{Bing Xu},
  \bibinfo{person}{David Warde-Farley}, \bibinfo{person}{Sherjil Ozair},
  \bibinfo{person}{Aaron Courville}, {and} \bibinfo{person}{Yoshua Bengio}.}
  \bibinfo{year}{2014}\natexlab{}.
\newblock \showarticletitle{Generative Adversarial Nets}. In
  \bibinfo{booktitle}{\emph{NeurIPS}}.
\newblock


\bibitem[Gu et~al\mbox{.}(2017)]%
        {gu2017badnets}
\bibfield{author}{\bibinfo{person}{Tianyu Gu}, \bibinfo{person}{Brendan
  Dolan-Gavitt}, {and} \bibinfo{person}{Siddharth Garg}.}
  \bibinfo{year}{2017}\natexlab{}.
\newblock \showarticletitle{Badnets: Identifying Vulnerabilities in the Machine
  Learning Model Supply Chain}.
\newblock \bibinfo{journal}{\emph{arXiv preprint arXiv:1708.06733}}
  (\bibinfo{year}{2017}).
\newblock


\bibitem[Guo et~al\mbox{.}(2019)]%
        {guo2019tabor}
\bibfield{author}{\bibinfo{person}{Wenbo Guo}, \bibinfo{person}{Lun Wang},
  \bibinfo{person}{Xinyu Xing}, \bibinfo{person}{Min Du}, {and}
  \bibinfo{person}{Dawn Song}.} \bibinfo{year}{2019}\natexlab{}.
\newblock \showarticletitle{Tabor: A Highly Accurate Approach to Inspecting and
  Restoring Trojan Backdoors in Ai Systems}.
\newblock \bibinfo{journal}{\emph{arXiv preprint arXiv:1908.01763}}
  (\bibinfo{year}{2019}).
\newblock


\bibitem[Hayase et~al\mbox{.}(2021)]%
        {hayase2021spectre}
\bibfield{author}{\bibinfo{person}{Jonathan Hayase}, \bibinfo{person}{Weihao
  Kong}, \bibinfo{person}{Raghav Somani}, {and} \bibinfo{person}{Sewoong Oh}.}
  \bibinfo{year}{2021}\natexlab{}.
\newblock \showarticletitle{SPECTRE: Defending Against Backdoor Attacks using
  Robust Statistics}. In \bibinfo{booktitle}{\emph{ICML}}.
\newblock


\bibitem[He et~al\mbox{.}(2016)]%
        {he2016deep}
\bibfield{author}{\bibinfo{person}{Kaiming He}, \bibinfo{person}{Xiangyu
  Zhang}, \bibinfo{person}{Shaoqing Ren}, {and} \bibinfo{person}{Jian Sun}.}
  \bibinfo{year}{2016}\natexlab{}.
\newblock \showarticletitle{Deep Residual Learning for Image Recognition}. In
  \bibinfo{booktitle}{\emph{CVPR}}.
\newblock


\bibitem[Hinton et~al\mbox{.}(2015)]%
        {hinton2015distilling}
\bibfield{author}{\bibinfo{person}{Geoffrey Hinton}, \bibinfo{person}{Oriol
  Vinyals}, {and} \bibinfo{person}{Jeff Dean}.}
  \bibinfo{year}{2015}\natexlab{}.
\newblock \showarticletitle{Distilling the Knowledge in a Neural Network}.
\newblock \bibinfo{journal}{\emph{arXiv preprint arXiv:1503.02531}}
  (\bibinfo{year}{2015}).
\newblock


\bibitem[Jia et~al\mbox{.}(2022)]%
        {jia2022badencoder}
\bibfield{author}{\bibinfo{person}{Jinyuan Jia}, \bibinfo{person}{Yupei Liu},
  {and} \bibinfo{person}{Neil~Zhenqiang Gong}.}
  \bibinfo{year}{2022}\natexlab{}.
\newblock \showarticletitle{Badencoder: Backdoor Attacks to Pre-trained
  Encoders in Self-supervised Learning}. In \bibinfo{booktitle}{\emph{2022 IEEE
  Symposium on Security and Privacy (SP)}}. IEEE, \bibinfo{pages}{2043--2059}.
\newblock


\bibitem[Kolouri et~al\mbox{.}(2019)]%
        {kolouri2019generalized}
\bibfield{author}{\bibinfo{person}{Soheil Kolouri}, \bibinfo{person}{Kimia
  Nadjahi}, \bibinfo{person}{Umut Simsekli}, \bibinfo{person}{Roland Badeau},
  {and} \bibinfo{person}{K Gustavo}.} \bibinfo{year}{2019}\natexlab{}.
\newblock \showarticletitle{Generalized Sliced Wasserstein Distances}. In
  \bibinfo{booktitle}{\emph{NeurIPS}}.
\newblock


\bibitem[Krizhevsky(2009)]%
        {CIFAR}
\bibfield{author}{\bibinfo{person}{Alex Krizhevsky}.}
  \bibinfo{year}{2009}\natexlab{}.
\newblock \showarticletitle{Learning Multiple Layers of Features from Tiny
  Images}.
\newblock \bibinfo{journal}{\emph{Master's thesis, University of Toronto}}
  (\bibinfo{year}{2009}).
\newblock


\bibitem[Li et~al\mbox{.}(2020a)]%
        {8892628}
\bibfield{author}{\bibinfo{person}{Gaolei Li}, \bibinfo{person}{Kaoru Ota},
  \bibinfo{person}{Mianxiong Dong}, \bibinfo{person}{Jun Wu}, {and}
  \bibinfo{person}{Jianhua Li}.} \bibinfo{year}{2020}\natexlab{a}.
\newblock \showarticletitle{DeSVig: Decentralized Swift Vigilance Against
  Adversarial Attacks in Industrial Artificial Intelligence Systems}.
\newblock \bibinfo{journal}{\emph{IEEE TII}} \bibinfo{volume}{16},
  \bibinfo{number}{5} (\bibinfo{year}{2020}), \bibinfo{pages}{3267--3277}.
\newblock


\bibitem[Li et~al\mbox{.}(2020c)]%
        {li2020invisible}
\bibfield{author}{\bibinfo{person}{Shaofeng Li}, \bibinfo{person}{Minhui Xue},
  \bibinfo{person}{Benjamin Zhao}, \bibinfo{person}{Haojin Zhu}, {and}
  \bibinfo{person}{Xinpeng Zhang}.} \bibinfo{year}{2020}\natexlab{c}.
\newblock \showarticletitle{Invisible Backdoor Attacks on Deep Neural Networks
  via Steganography and Regularization}.
\newblock \bibinfo{journal}{\emph{IEEE TDSC}} (\bibinfo{year}{2020}).
\newblock


\bibitem[Li et~al\mbox{.}(2021a)]%
        {li2021invisible}
\bibfield{author}{\bibinfo{person}{Yuezun Li}, \bibinfo{person}{Yiming Li},
  \bibinfo{person}{Baoyuan Wu}, \bibinfo{person}{Longkang Li},
  \bibinfo{person}{Ran He}, {and} \bibinfo{person}{Siwei Lyu}.}
  \bibinfo{year}{2021}\natexlab{a}.
\newblock \showarticletitle{Invisible Backdoor Attack with Sample-specific
  Triggers}. In \bibinfo{booktitle}{\emph{ICCV}}.
  \bibinfo{pages}{16463--16472}.
\newblock


\bibitem[Li et~al\mbox{.}(2021b)]%
        {li2021anti}
\bibfield{author}{\bibinfo{person}{Yige Li}, \bibinfo{person}{Xixiang Lyu},
  \bibinfo{person}{Nodens Koren}, \bibinfo{person}{Lingjuan Lyu},
  \bibinfo{person}{Bo Li}, {and} \bibinfo{person}{Xingjun Ma}.}
  \bibinfo{year}{2021}\natexlab{b}.
\newblock \showarticletitle{Anti-backdoor Learning: Training Clean Models on
  Poisoned Data}. In \bibinfo{booktitle}{\emph{NeurIPS}}.
\newblock


\bibitem[Li et~al\mbox{.}(2021c)]%
        {li2021neural}
\bibfield{author}{\bibinfo{person}{Yige Li}, \bibinfo{person}{Xixiang Lyu},
  \bibinfo{person}{Nodens Koren}, \bibinfo{person}{Lingjuan Lyu},
  \bibinfo{person}{Bo Li}, {and} \bibinfo{person}{Xingjun Ma}.}
  \bibinfo{year}{2021}\natexlab{c}.
\newblock \showarticletitle{Neural Attention Distillation: Erasing Backdoor
  Triggers from Deep Neural Networks}. In \bibinfo{booktitle}{\emph{ICLR}}.
\newblock


\bibitem[Li et~al\mbox{.}(2020b)]%
        {li2020backdoor}
\bibfield{author}{\bibinfo{person}{Yiming Li}, \bibinfo{person}{Baoyuan Wu},
  \bibinfo{person}{Yong Jiang}, \bibinfo{person}{Zhifeng Li}, {and}
  \bibinfo{person}{Shu-Tao Xia}.} \bibinfo{year}{2020}\natexlab{b}.
\newblock \showarticletitle{Backdoor Learning: A Survey}.
\newblock \bibinfo{journal}{\emph{arXiv preprint arXiv:2007.08745}}
  (\bibinfo{year}{2020}).
\newblock


\bibitem[Lin et~al\mbox{.}(2020)]%
        {lin2020composite}
\bibfield{author}{\bibinfo{person}{Junyu Lin}, \bibinfo{person}{Lei Xu},
  \bibinfo{person}{Yingqi Liu}, {and} \bibinfo{person}{Xiangyu Zhang}.}
  \bibinfo{year}{2020}\natexlab{}.
\newblock \showarticletitle{Composite Backdoor Attack for Deep Neural Network
  by Mixing Existing Benign Features}. In \bibinfo{booktitle}{\emph{CCS}}.
  \bibinfo{pages}{113--131}.
\newblock


\bibitem[Liu et~al\mbox{.}(2018a)]%
        {liu2018fine}
\bibfield{author}{\bibinfo{person}{Kang Liu}, \bibinfo{person}{Brendan
  Dolan-Gavitt}, {and} \bibinfo{person}{Siddharth Garg}.}
  \bibinfo{year}{2018}\natexlab{a}.
\newblock \showarticletitle{Fine-pruning: Defending against Backdooring Attacks
  on Deep Neural Networks}. In \bibinfo{booktitle}{\emph{RAID}}.
\newblock


\bibitem[Liu et~al\mbox{.}(2021a)]%
        {liu2021removing}
\bibfield{author}{\bibinfo{person}{Xuankai Liu}, \bibinfo{person}{Fengting Li},
  \bibinfo{person}{Bihan Wen}, {and} \bibinfo{person}{Qi Li}.}
  \bibinfo{year}{2021}\natexlab{a}.
\newblock \showarticletitle{Removing Backdoor-based Watermarks in Neural
  Networks with Limited Data}. In \bibinfo{booktitle}{\emph{ICPR}}.
\newblock


\bibitem[Liu et~al\mbox{.}(2019)]%
        {liu2019abs}
\bibfield{author}{\bibinfo{person}{Yingqi Liu}, \bibinfo{person}{Wen-Chuan
  Lee}, \bibinfo{person}{Guanhong Tao}, \bibinfo{person}{Shiqing Ma},
  \bibinfo{person}{Yousra Aafer}, {and} \bibinfo{person}{Xiangyu Zhang}.}
  \bibinfo{year}{2019}\natexlab{}.
\newblock \showarticletitle{Abs: Scanning Neural Networks for Back-doors by
  Artificial Brain Stimulation}. In \bibinfo{booktitle}{\emph{CCS}}.
  \bibinfo{pages}{1265--1282}.
\newblock


\bibitem[Liu et~al\mbox{.}(2018b)]%
        {liu2017trojaning}
\bibfield{author}{\bibinfo{person}{Yingqi Liu}, \bibinfo{person}{Shiqing Ma},
  \bibinfo{person}{Yousra Aafer}, \bibinfo{person}{Wen{-}Chuan Lee},
  \bibinfo{person}{Juan Zhai}, \bibinfo{person}{Weihang Wang}, {and}
  \bibinfo{person}{Xiangyu Zhang}.} \bibinfo{year}{2018}\natexlab{b}.
\newblock \showarticletitle{Trojaning Attack on Neural Networks}. In
  \bibinfo{booktitle}{\emph{NDSS}}.
\newblock


\bibitem[Liu et~al\mbox{.}(2020)]%
        {liu2020reflection}
\bibfield{author}{\bibinfo{person}{Yunfei Liu}, \bibinfo{person}{Xingjun Ma},
  \bibinfo{person}{James Bailey}, {and} \bibinfo{person}{Feng Lu}.}
  \bibinfo{year}{2020}\natexlab{}.
\newblock \showarticletitle{Reflection Backdoor: A Natural Backdoor Attack on
  Deep Neural Networks}. In \bibinfo{booktitle}{\emph{ECCV}}.
\newblock


\bibitem[Liu et~al\mbox{.}(2021b)]%
        {liu2021data}
\bibfield{author}{\bibinfo{person}{Yuang Liu}, \bibinfo{person}{Wei Zhang},
  \bibinfo{person}{Jun Wang}, {and} \bibinfo{person}{Jianyong Wang}.}
  \bibinfo{year}{2021}\natexlab{b}.
\newblock \showarticletitle{Data-free Knowledge Transfer: A Survey}.
\newblock \bibinfo{journal}{\emph{arXiv preprint arXiv:2112.15278}}
  (\bibinfo{year}{2021}).
\newblock


\bibitem[Paszke et~al\mbox{.}(2017)]%
        {paszke2017automatic}
\bibfield{author}{\bibinfo{person}{Adam Paszke}, \bibinfo{person}{Sam Gross},
  \bibinfo{person}{Soumith Chintala}, \bibinfo{person}{Gregory Chanan},
  \bibinfo{person}{Edward Yang}, \bibinfo{person}{Zachary DeVito},
  \bibinfo{person}{Zeming Lin}, \bibinfo{person}{Alban Desmaison},
  \bibinfo{person}{Luca Antiga}, {and} \bibinfo{person}{Adam Lerer}.}
  \bibinfo{year}{2017}\natexlab{}.
\newblock \showarticletitle{Automatic Differentiation in PyTorch}. In
  \bibinfo{booktitle}{\emph{NIPS-W}}.
\newblock


\bibitem[Qiao et~al\mbox{.}(2019)]%
        {qiao2019defending}
\bibfield{author}{\bibinfo{person}{Ximing Qiao}, \bibinfo{person}{Yukun Yang},
  {and} \bibinfo{person}{Hai Li}.} \bibinfo{year}{2019}\natexlab{}.
\newblock \showarticletitle{Defending Neural Backdoors via Generative
  Distribution Modeling}. In \bibinfo{booktitle}{\emph{NeurIPS}}.
\newblock


\bibitem[Ren et~al\mbox{.}(2022)]%
        {9954194}
\bibfield{author}{\bibinfo{person}{Ge Ren}, \bibinfo{person}{Jun Wu},
  \bibinfo{person}{Gaolei Li}, \bibinfo{person}{Shenghong Li}, {and}
  \bibinfo{person}{Mohsen Guizani}.} \bibinfo{year}{2022}\natexlab{}.
\newblock \showarticletitle{Protecting Intellectual Property with Reliable
  Availability of Learning Models in AI-based Cybersecurity Services}.
\newblock \bibinfo{journal}{\emph{IEEE TDSC}} (\bibinfo{year}{2022}),
  \bibinfo{pages}{1--18}.
\newblock


\bibitem[Saha et~al\mbox{.}(2020)]%
        {saha2020hidden}
\bibfield{author}{\bibinfo{person}{Aniruddha Saha},
  \bibinfo{person}{Akshayvarun Subramanya}, {and} \bibinfo{person}{Hamed
  Pirsiavash}.} \bibinfo{year}{2020}\natexlab{}.
\newblock \showarticletitle{Hidden Trigger Backdoor Attacks}. In
  \bibinfo{booktitle}{\emph{AAAI}}.
\newblock


\bibitem[Tian et~al\mbox{.}(2020)]%
        {tian2020self}
\bibfield{author}{\bibinfo{person}{Yuan Tian}, \bibinfo{person}{Zhaohui Che},
  \bibinfo{person}{Wenbo Bao}, \bibinfo{person}{Guangtao Zhai}, {and}
  \bibinfo{person}{Zhiyong Gao}.} \bibinfo{year}{2020}\natexlab{}.
\newblock \showarticletitle{Self-supervised motion representation via
  scattering local motion cues}. In \bibinfo{booktitle}{\emph{European
  Conference on Computer Vision}}. Springer, \bibinfo{pages}{71--89}.
\newblock


\bibitem[Tian et~al\mbox{.}(2021)]%
        {tian2021self}
\bibfield{author}{\bibinfo{person}{Yuan Tian}, \bibinfo{person}{Guo Lu},
  \bibinfo{person}{Xiongkuo Min}, \bibinfo{person}{Zhaohui Che},
  \bibinfo{person}{Guangtao Zhai}, \bibinfo{person}{Guodong Guo}, {and}
  \bibinfo{person}{Zhiyong Gao}.} \bibinfo{year}{2021}\natexlab{}.
\newblock \showarticletitle{Self-conditioned probabilistic learning of video
  rescaling}. In \bibinfo{booktitle}{\emph{CVPR}}. \bibinfo{pages}{4490--4499}.
\newblock


\bibitem[Tian et~al\mbox{.}(2022)]%
        {tian2022ean}
\bibfield{author}{\bibinfo{person}{Yuan Tian}, \bibinfo{person}{Yichao Yan},
  \bibinfo{person}{Guangtao Zhai}, \bibinfo{person}{Guodong Guo}, {and}
  \bibinfo{person}{Zhiyong Gao}.} \bibinfo{year}{2022}\natexlab{}.
\newblock \showarticletitle{Ean: event adaptive network for enhanced action
  recognition}.
\newblock \bibinfo{journal}{\emph{IJCV}} \bibinfo{volume}{130},
  \bibinfo{number}{10} (\bibinfo{year}{2022}), \bibinfo{pages}{2453--2471}.
\newblock


\bibitem[Tran et~al\mbox{.}(2018)]%
        {tran2018spectral}
\bibfield{author}{\bibinfo{person}{Brandon Tran}, \bibinfo{person}{Jerry Li},
  {and} \bibinfo{person}{Aleksander Madry}.} \bibinfo{year}{2018}\natexlab{}.
\newblock \showarticletitle{Spectral Signatures in Backdoor Attacks}. In
  \bibinfo{booktitle}{\emph{NeurIPS}}.
\newblock


\bibitem[Turner et~al\mbox{.}(2019)]%
        {turner2019label}
\bibfield{author}{\bibinfo{person}{Alexander Turner}, \bibinfo{person}{Dimitris
  Tsipras}, {and} \bibinfo{person}{Aleksander Madry}.}
  \bibinfo{year}{2019}\natexlab{}.
\newblock \showarticletitle{Label-Consistent Backdoor Attacks}.
\newblock \bibinfo{journal}{\emph{arXiv preprint arXiv:1912.02771}}
  (\bibinfo{year}{2019}).
\newblock


\bibitem[Villarreal-Vasquez and Bhargava(2020)]%
        {villarreal2020confoc}
\bibfield{author}{\bibinfo{person}{Miguel Villarreal-Vasquez} {and}
  \bibinfo{person}{Bharat Bhargava}.} \bibinfo{year}{2020}\natexlab{}.
\newblock \showarticletitle{Confoc: Content-focus Protection against Trojan
  Attacks on Neural Networks}.
\newblock \bibinfo{journal}{\emph{arXiv preprint arXiv:2007.00711}}
  (\bibinfo{year}{2020}).
\newblock


\bibitem[Vinyals et~al\mbox{.}(2016)]%
        {vinyals2016matching}
\bibfield{author}{\bibinfo{person}{Oriol Vinyals}, \bibinfo{person}{Charles
  Blundell}, \bibinfo{person}{Timothy Lillicrap}, \bibinfo{person}{koray
  kavukcuoglu}, {and} \bibinfo{person}{Daan Wierstra}.}
  \bibinfo{year}{2016}\natexlab{}.
\newblock \showarticletitle{Matching Networks for One Shot Learning}. In
  \bibinfo{booktitle}{\emph{NeurIPS}}.
\newblock


\bibitem[Wang et~al\mbox{.}(2019)]%
        {wang2019neural}
\bibfield{author}{\bibinfo{person}{Bolun Wang}, \bibinfo{person}{Yuanshun Yao},
  \bibinfo{person}{Shawn Shan}, \bibinfo{person}{Huiying Li},
  \bibinfo{person}{Bimal Viswanath}, \bibinfo{person}{Haitao Zheng}, {and}
  \bibinfo{person}{Ben~Y Zhao}.} \bibinfo{year}{2019}\natexlab{}.
\newblock \showarticletitle{Neural Cleanse: Identifying and Mitigating Backdoor
  Attacks in Neural Networks}. In \bibinfo{booktitle}{\emph{SP}}.
\newblock


\bibitem[Wang et~al\mbox{.}(2020)]%
        {wang2020backdoor}
\bibfield{author}{\bibinfo{person}{Shuo Wang}, \bibinfo{person}{Surya Nepal},
  \bibinfo{person}{Carsten Rudolph}, \bibinfo{person}{Marthie Grobler},
  \bibinfo{person}{Shangyu Chen}, {and} \bibinfo{person}{Tianle Chen}.}
  \bibinfo{year}{2020}\natexlab{}.
\newblock \showarticletitle{Backdoor Attacks against Transfer Learning with
  Pre-trained Deep Learning Models}.
\newblock \bibinfo{journal}{\emph{IEEE TSC}} (\bibinfo{year}{2020}).
\newblock


\bibitem[Wang et~al\mbox{.}(2018)]%
        {wang2018zero}
\bibfield{author}{\bibinfo{person}{Xiaolong Wang}, \bibinfo{person}{Yufei Ye},
  {and} \bibinfo{person}{Abhinav Gupta}.} \bibinfo{year}{2018}\natexlab{}.
\newblock \showarticletitle{Zero-shot Recognition via Semantic Embeddings and
  Knowledge Graphs}. In \bibinfo{booktitle}{\emph{CVPR}}.
  \bibinfo{pages}{6857--6866}.
\newblock


\bibitem[Xie et~al\mbox{.}(2019)]%
        {xie2019dba}
\bibfield{author}{\bibinfo{person}{Chulin Xie}, \bibinfo{person}{Keli Huang},
  \bibinfo{person}{Pin-Yu Chen}, {and} \bibinfo{person}{Bo Li}.}
  \bibinfo{year}{2019}\natexlab{}.
\newblock \showarticletitle{DBA: Distributed Backdoor Attacks against Federated
  Learning}. In \bibinfo{booktitle}{\emph{ICLR}}.
\newblock


\bibitem[Xu et~al\mbox{.}(2021)]%
        {xu2021detecting}
\bibfield{author}{\bibinfo{person}{Xiaojun Xu}, \bibinfo{person}{Qi Wang},
  \bibinfo{person}{Huichen Li}, \bibinfo{person}{Nikita Borisov},
  \bibinfo{person}{Carl~A Gunter}, {and} \bibinfo{person}{Bo Li}.}
  \bibinfo{year}{2021}\natexlab{}.
\newblock \showarticletitle{Detecting AI Trojans using Meta Neural Analysis}.
  In \bibinfo{booktitle}{\emph{SP}}. \bibinfo{pages}{103--120}.
\newblock


\bibitem[Yan et~al\mbox{.}(2021a)]%
        {yan2021dehib}
\bibfield{author}{\bibinfo{person}{Zhicong Yan}, \bibinfo{person}{Gaolei Li},
  \bibinfo{person}{Yuan TIan}, \bibinfo{person}{Jun Wu},
  \bibinfo{person}{Shenghong Li}, \bibinfo{person}{Mingzhe Chen}, {and}
  \bibinfo{person}{H~Vincent Poor}.} \bibinfo{year}{2021}\natexlab{a}.
\newblock \showarticletitle{Dehib: Deep Hidden Backdoor Attack on
  Semi-supervised Learning via Adversarial Perturbation}. In
  \bibinfo{booktitle}{\emph{AAAI}}, Vol.~\bibinfo{volume}{35}.
  \bibinfo{pages}{10585--10593}.
\newblock


\bibitem[Yan et~al\mbox{.}(2021b)]%
        {yan2021deep}
\bibfield{author}{\bibinfo{person}{Zhicong Yan}, \bibinfo{person}{Jun Wu},
  \bibinfo{person}{Gaolei Li}, \bibinfo{person}{Shenghong Li}, {and}
  \bibinfo{person}{Mohsen Guizani}.} \bibinfo{year}{2021}\natexlab{b}.
\newblock \showarticletitle{Deep Neural Backdoor in Semi-Supervised Learning:
  Threats and Countermeasures}.
\newblock \bibinfo{journal}{\emph{IEEE TIFS}}  \bibinfo{volume}{16}
  (\bibinfo{year}{2021}), \bibinfo{pages}{4827--4842}.
\newblock


\bibitem[Yao et~al\mbox{.}(2019)]%
        {yao2019latent}
\bibfield{author}{\bibinfo{person}{Yuanshun Yao}, \bibinfo{person}{Huiying Li},
  \bibinfo{person}{Haitao Zheng}, {and} \bibinfo{person}{Ben~Y Zhao}.}
  \bibinfo{year}{2019}\natexlab{}.
\newblock \showarticletitle{Latent Backdoor Attacks on Deep Neural Networks}.
  In \bibinfo{booktitle}{\emph{CCS}}. \bibinfo{pages}{2041--2055}.
\newblock


\bibitem[Yin et~al\mbox{.}(2021)]%
        {yin2021see}
\bibfield{author}{\bibinfo{person}{Hongxu Yin}, \bibinfo{person}{Arun Mallya},
  \bibinfo{person}{Arash Vahdat}, \bibinfo{person}{Jose~M Alvarez},
  \bibinfo{person}{Jan Kautz}, {and} \bibinfo{person}{Pavlo Molchanov}.}
  \bibinfo{year}{2021}\natexlab{}.
\newblock \showarticletitle{See Through Gradients: Image Batch Recovery via
  Gradinversion}. In \bibinfo{booktitle}{\emph{CVPR}}.
  \bibinfo{pages}{16337--16346}.
\newblock


\bibitem[Yin et~al\mbox{.}(2020)]%
        {yin2020dreaming}
\bibfield{author}{\bibinfo{person}{Hongxu Yin}, \bibinfo{person}{Pavlo
  Molchanov}, \bibinfo{person}{Jose~M Alvarez}, \bibinfo{person}{Zhizhong Li},
  \bibinfo{person}{Arun Mallya}, \bibinfo{person}{Derek Hoiem},
  \bibinfo{person}{Niraj~K Jha}, {and} \bibinfo{person}{Jan Kautz}.}
  \bibinfo{year}{2020}\natexlab{}.
\newblock \showarticletitle{Dreaming to Distill: Data-free Knowledge Transfer
  via Deepinversion}. In \bibinfo{booktitle}{\emph{CVPR}}.
  \bibinfo{pages}{8715--8724}.
\newblock


\bibitem[Zhao et~al\mbox{.}(2019)]%
        {zhao2019bridging}
\bibfield{author}{\bibinfo{person}{Pu Zhao}, \bibinfo{person}{Pin-Yu Chen},
  \bibinfo{person}{Payel Das}, \bibinfo{person}{Karthikeyan~Natesan
  Ramamurthy}, {and} \bibinfo{person}{Xue Lin}.}
  \bibinfo{year}{2019}\natexlab{}.
\newblock \showarticletitle{Bridging Mode Connectivity in Loss Landscapes and
  Adversarial Robustness}. In \bibinfo{booktitle}{\emph{ICLR}}.
\newblock


\end{thebibliography}

\appendix

\section*{Appendix}

\section{More Attack Implementation Details}

Detailed implementation of backdoor attacks are summarized as follows:

\textbf{(a) Badnets:} We perform Badnets attack by poisoning the training dataset, where we randomly select $N_{poison}$ training images from the whole dataset (except images from the target class $t$), injecting triggers into them and flipping their labels to the target class $t$. Then the victim model is trained on the poisoned dataset using an SGD optimizer with initial learning rate of 0.1, momentum of 0.9, and weight decay of 5e-4. $N_{poison} = 300$ is enough to achieve a high attack success rate and we use this setting for all Badnets experiments. We finetune the model for 200 epochs with learning rate decay of 0.1 at epoch 100 and epoch 150. The training data augmentations are limited to Random Crop and Random Horizontal Flip only. In the experiments section, we always take the last class in the dataset as the target class (i.e. "truck" in CIFAR10 and "tractor" in CIFAR100). The attack and defense results of other classes is listed in Table \ref{tab:results_cifar_other_classes}.

\textbf{(b) Clean-Label:} 
Clean-label attack poisons the dataset without flipping the training labels, but it corrupts the salient object feature in the poisoned images and forces the model's decisions to be more dependent on the trigger pattern \cite{turner2019label}.
In Clean-label attacks, we randomly choose 300 training samples from the target category, use adversarial perturbations to corrupt their salient object feature, and stamp square triggers to them at the right-bottom location. Then we perform the normal learning procedure with the same training settings as the Badnets attack.


\textbf{(c) Trojaning:} In this work, we first randomly select 10 neurons of the $Average Pooling$ layer in Resnet-18 model (shown in Table \ref{tab:resnet_arch}), reverse engineers a trigger pattern that can achieve maximum responses of selected neurons, then finetunes only the subsequent $FC$ layer to strength the connection between the selected neurons and the target class score using a poisoned dataset with $N_{poison}=300$ by inserting the reverse engineered trigger into them.

\section{Model Architectures and Hyperparameters}

\subsection{Architecture of Backdoored Teachers and Students}

We list the employed architecture of Resnet-18 in Table \ref{tab:resnet_arch}, and use it as a unified model architecture for fair comparisons of various backdoor erasing methods.
During the training or finetuning process of the Rsenet-18, we perform a normalization operation after randomized preprocessing on each sample according to the statistics of the employed dataset, so the model is fed with data that has zero means and unit variances.

\subsection{Architecture of Sample Generators and Trigger Generators}

We also list the architectures of employed sample generators $\mathcal{G}$ and trigger generators $\mathcal{G}_p$ that produce different size of outputs in Table \ref{tab:trigger_generator1}, \ref{tab:trigger_generator2}, \ref{tab:trigger_generator3} and \ref{tab:sample_generator}.
Specifically, the sample generators are ended with batch normalization layers, so that the produced samples should have zero means and unit variances, and can be directly fed into the teacher network and student network to perform knowledge distillation.
The trigger generators are ended with $Tanh$ layers, so that their outputs are clamped within $(-1,1)$. The slope of $LeakyReLU$ is set to 0.2 and all generators contains $BN$ layers for stabilized training.

\subsection{Mixing Generated Triggers to Fake Samples}
In every step where a produced trigger $\delta$ is injected to a produced fake sample $x$, we first denormalize the fake sample, then add the produced trigger $\delta$ to the denormalized sample, and finally perform the normalization process.

\begin{table}[htb!]
    \centering
    \caption{DHBE results against Badnets attack on other classes of CIFAR10 ($3\times3$ square trigger is used). Numbers are displayed as percentages. }
    \begin{tabular}{c|cc|cc}
\toprule
\multirow{2}{*}{Target Class} & \multicolumn{2}{c|} {Backdoored} & \multicolumn{2}{c} {DHBE} \\
& ACC & ASR & ACC & ASR \\
\hline
airplane & 95.10 & 100.00 & 94.55 & 2.74 \\
automobile & 95.09 & 99.98 & 94.46 & 0.30 \\
bird & 94.83 & 100.00 & 94.16 & 0.15 \\
cat & 95.14 & 100.00 & 94.54 & 1.22 \\
deer & 94.74 & 100.00 & 93.80 & 0.14 \\
dog & 94.87 & 100.00 & 94.18 & 0.18 \\
frog & 94.69 & 100.00 & 94.59 & 0.66 \\
horse & 95.18 & 100.00 & 94.48 & 0.01 \\
ship & 94.70 & 100.00 & 94.21 & 0.08 \\
\bottomrule
    \end{tabular}
    \label{tab:results_cifar_other_classes}
\end{table}

\begin{table}[htb!]
    \centering
    \caption{DHBE results against Badnets attack on other network architectures ($3\times3$ square trigger is used). Numbers are displayed as percentages. }
    \begin{tabular}{c|cc|cc}
\toprule
\multirow{2}{*}{Target Class} & \multicolumn{2}{c|} {Backdoored} & \multicolumn{2}{c} {DHBE} \\
& ACC & ASR & ACC & ASR \\
\hline
Resnet34 & 95.53 & 100.00 & 94.50 & 0.26 \\
WRN40-1 & 92.99 & 99.99 & 92.50 & 0.45 \\
WRN40-2 & 94.94 & 99.96 & 94.44 & 1.42 \\
\bottomrule
    \end{tabular}
    \label{tab:results_cifar_other_networks}
\end{table}

\begin{table}[htb!]
    \centering
    \caption{Resnet-18 architectures used in this work.}
    \begin{tabular}{c|c}
    \toprule
        Resnet-18  & Output Size  \\
        \hline 
        Input & $32\times 32 \times 3$ / $64 \times 64 \times 3$ \\
        \hline
        $3\times3\;64\;S1\;Conv$  / $5\times5\;64\;S2\;Conv$ & 
            $32\times 32 \times 64$ \\
        \hline
        $BN, ReLU$ & $32\times 32 \times 64$ \\
        \hline
        \multirow{3}{*}{$\begin{bmatrix} 3\times3\;64\;Conv,BN,ReLU \\ 3\times3\;64\;Conv, BN, ReLU \end{bmatrix} \times 2$} & 
            \multirow{3}{*}{$32\times 32 \times 64$} \\
         & \\
         & \\
        \hline
        \multirow{3}{*}{$\begin{bmatrix} 3\times3\;128\;Conv,BN,ReLU \\ 3\times3\;128\;Conv, BN, ReLU \end{bmatrix} \times 2$} & 
            \multirow{3}{*}{$16\times 16 \times 128$} \\
         & \\
         & \\
        \hline
        \multirow{3}{*}{$\begin{bmatrix} 3\times3\;256\;Conv,BN,ReLU \\ 3\times3\;256\;Conv, BN, ReLU \end{bmatrix} \times 2$} & 
            \multirow{3}{*}{$8\times 8 \times 256$} \\
         & \\
         & \\
        \hline
        \multirow{3}{*}{$\begin{bmatrix} 3\times3\;512\;Conv,BN,ReLU \\ 3\times3\;512\;Conv, BN, ReLU \end{bmatrix} \times 2$} & 
            \multirow{3}{*}{$4\times 4 \times 512$} \\
         & \\
         & \\
         \hline
         Avg Pool & $ 512 $\\
         \hline
         FC & Num. of Classes\\
         \bottomrule
    \end{tabular}
    \label{tab:resnet_arch}
\end{table}

\begin{table}[htb!]
    \centering
    \caption{Architectures of trigger generators $\mathcal{G}_p$ of output size $3\times3$, $5\times5$ and $7\times7$.}
    \begin{tabular}{c|c|c|c}
        \toprule
        Trigger Generator $\mathcal{G}_p$ & \multicolumn{3}{c}{Output Size} \\
        \hline 
        Input Dim. of $z$ & \multicolumn{3}{c}{256} \\
        \hline
        $FC,Reshape,BN$ & 
            $3\times3\times64$ & 
                $5\times5\times64$ & 
                    $7\times7\times64$ \\
        \hline
        $3\times3\;3\;Conv, BN, Tanh$ & 
            $3\times3\times3$ & 
                $5\times5\times3$ & 
                    $7\times7\times3$ \\
        \bottomrule
    \end{tabular}
    \label{tab:trigger_generator1}
\end{table}

\begin{table}[h]
    \centering
    \caption{Architectures of trigger generators $\mathcal{G}_p$ of output size $10\times10$ and $14\times14$.}
    \begin{tabular}{c|c|c}
        \toprule
        Trigger Generator $\mathcal{G}_p$ & \multicolumn{2}{c}{Output Size} \\
        \hline 
        Input Dim. of $z$ & \multicolumn{2}{c}{256} \\
        \hline
        $FC,Reshape,BN$ & 
                $5\times5\times128$ & 
                    $7\times7\times128$ \\
        \hline
        $Upsample\;2\times$ & 
                $10\times10\times128$ & 
                    $14\times14\times128$ \\
        \hline
        $3\times3\;64\;Conv$ & 
                \multirow{2}{*}{$10\times10\times64$} & 
                    \multirow{2}{*}{$14\times14\times64$} \\
        $BN, LeakyReLU\;0.2$ &  &  \\
        \hline 
        $3\times3\;3\;Conv, BN, Tanh$ & 
                $10\times10\times3$ & 
                    $14\times14\times3$ \\
        \bottomrule
    \end{tabular}
    \label{tab:trigger_generator2}
\end{table}

\begin{table}[htb!]
    \centering
    \caption{Architectures of trigger generators $\mathcal{G}_p$ of output size $32\times32$.}
    \begin{tabular}{c|c}
        \toprule
        Trigger Generator $\mathcal{G}_p$ & Output Size \\
        \hline 
        Input Dim. of $z$ & 256 \\
        \hline
        $FC,Reshape,BN$ & 
                        $8\times8\times128$ \\
        \hline
        $Upsample\;2\times$ & 
                        $16\times16\times128$\\
        \hline
        $3\times3\;128\;Conv$ & 
                        \multirow{2}{*}{$16\times16\times128$}\\
        $BN, LeakyReLU\;0.2$  & \\
        \hline
        $Upsample\;2\times$ & 
                        $32\times32\times128$\\
        \hline
        $3\times3\;64\;Conv$ & 
                        \multirow{2}{*}{$32\times32\times64$}\\
        $BN, LeakyReLU\;0.2$ & \\
        \hline 
        $3\times3\;3\;Conv, BN, Tanh$ & 
                        $32\times32\times3$\\
        \bottomrule
    \end{tabular}
    \label{tab:trigger_generator3}
\end{table}

\begin{table}[h]
    \centering
    \caption{Architectures of sample generators $\mathcal{G}$ of different output size.}
    \begin{tabular}{c|c|c}
        \toprule
        Sample Generator $\mathcal{G}$ & \multicolumn{2}{c}{Output Size} \\
        \hline 
        Input Dim. of $z$ & \multicolumn{2}{c}{256} \\
        \hline
        $FC,Reshape,BN$ & 
            \multicolumn{2}{c}{$8\times8\times128$} \\
        \hline
        $Upsample\;2\times$ & 
            \multicolumn{2}{c}{$16\times16\times128$} \\
        \hline
        $3\times3\;128\;Conv$ & 
        
            \multicolumn{2}{c}{\multirow{2}{*}{$16\times16\times128$}} \\
        $BN, LeakyReLU\;0.2$ & 
             \multicolumn{2}{c}{} \\
             
        \hline
        $Upsample\;2\times$ & 
            \multicolumn{2}{c}{$32\times32\times128$} \\
        \hline
        
        $3\times3\;128\;Conv$ & 
            \multirow{2}{*}{-}  &
                \multirow{2}{*}{$32\times32\times128$}\\
        $BN, LeakyReLU\;0.2$ & 
             \\
        \hline
        $Upsample\;2\times$ & 
            - & 
                $64\times64\times128$ \\
        \hline 
        $3\times3\;64\;Conv$ & 
            \multirow{2}{*}{$32\times32\times64$}  &
                \multirow{2}{*}{$64\times64\times64$}\\
        $BN, LeakyReLU\;0.2$ & 
             \\
        \hline
        $3\times3\;3\;Conv, Sigmoid, BN$ & 
                $32\times32\times3$ & 
                    $64\times64\times3$ \\
        \bottomrule
    \end{tabular}
    \label{tab:sample_generator}
\end{table}

\section{Detailed Configuration of Blind Methods}

The detailed configuration and training process of blind methods is summarized as follows:

\textbf{(a) Finetuning, Finepruning \cite{liu2018fine}, and Neural Attention Distillation (NAD) \cite{li2021neural}.} For these methods, we adapt the same optimizer and learning rate settings: 
An SGD optimizer with initial learning rate of 0.01, momentum of 0.9 is employed. the backdoored model is fully finetuned by 20 epochs and 2000 samples are loaded for each epoch with batch size of 64. When the number of input training samples is less than 2000, we duplicate them until there are 2000 samples. The learning rate is decayed by 0.1 at epoch 10 and epoch 15.
For NAD, the attention transfer loss implemented in open-source code \footnote{\url{https://github.com/bboylyg/NAD}} is used. Between the teacher and student models, we add four attention transfer losses between the outputs of their four residual blocks, and each loss is multiplied by $beta=5000$ as suggested in \cite{li2021neural} before finally adding up with the cross-entropy loss and performing back-propagation optimization.

\begin{figure*}[htb!]
    \centering
    \begin{tabular}{ccc}
    \includegraphics[width=0.30\textwidth]{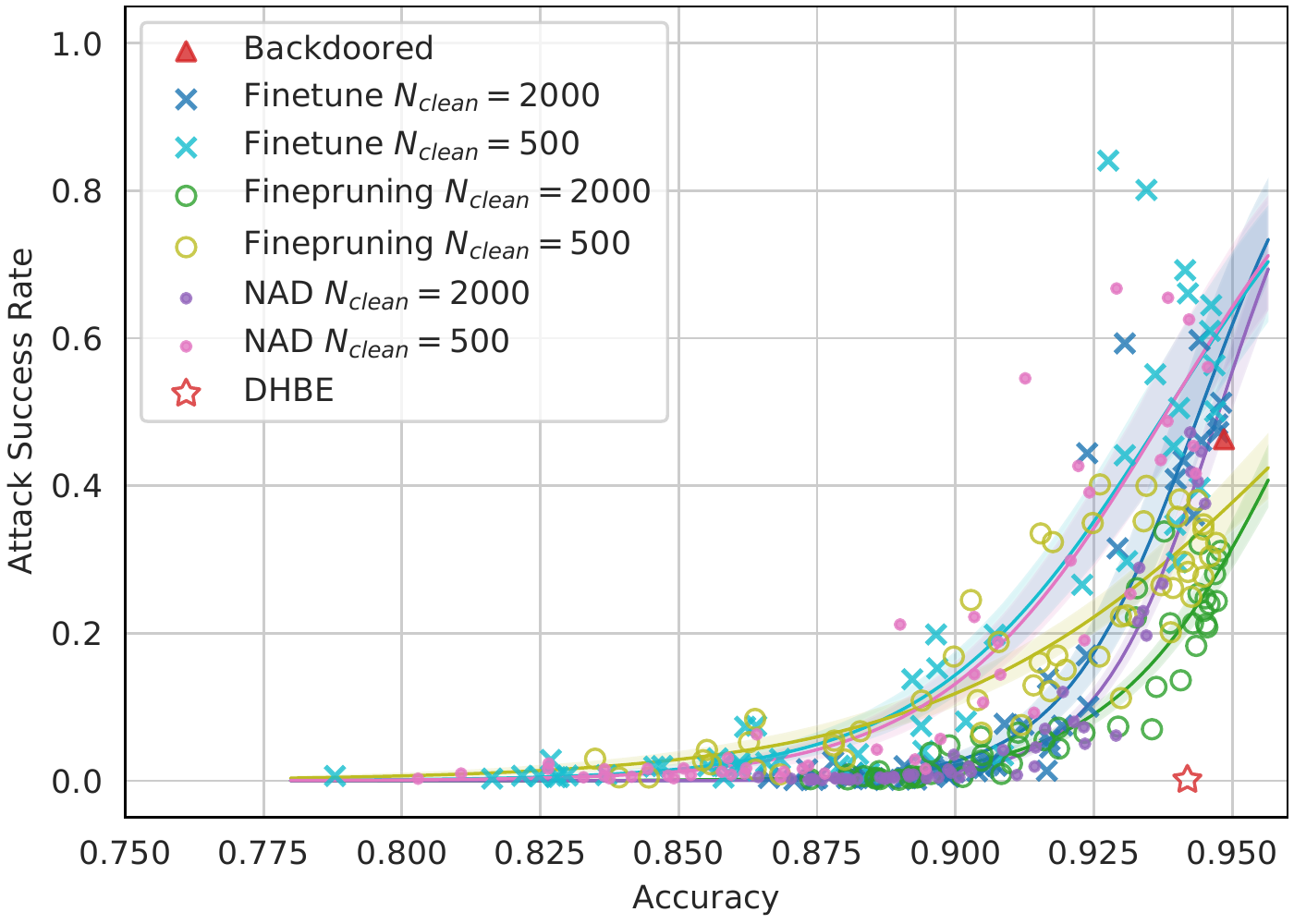}&
    \includegraphics[width=0.30\textwidth]{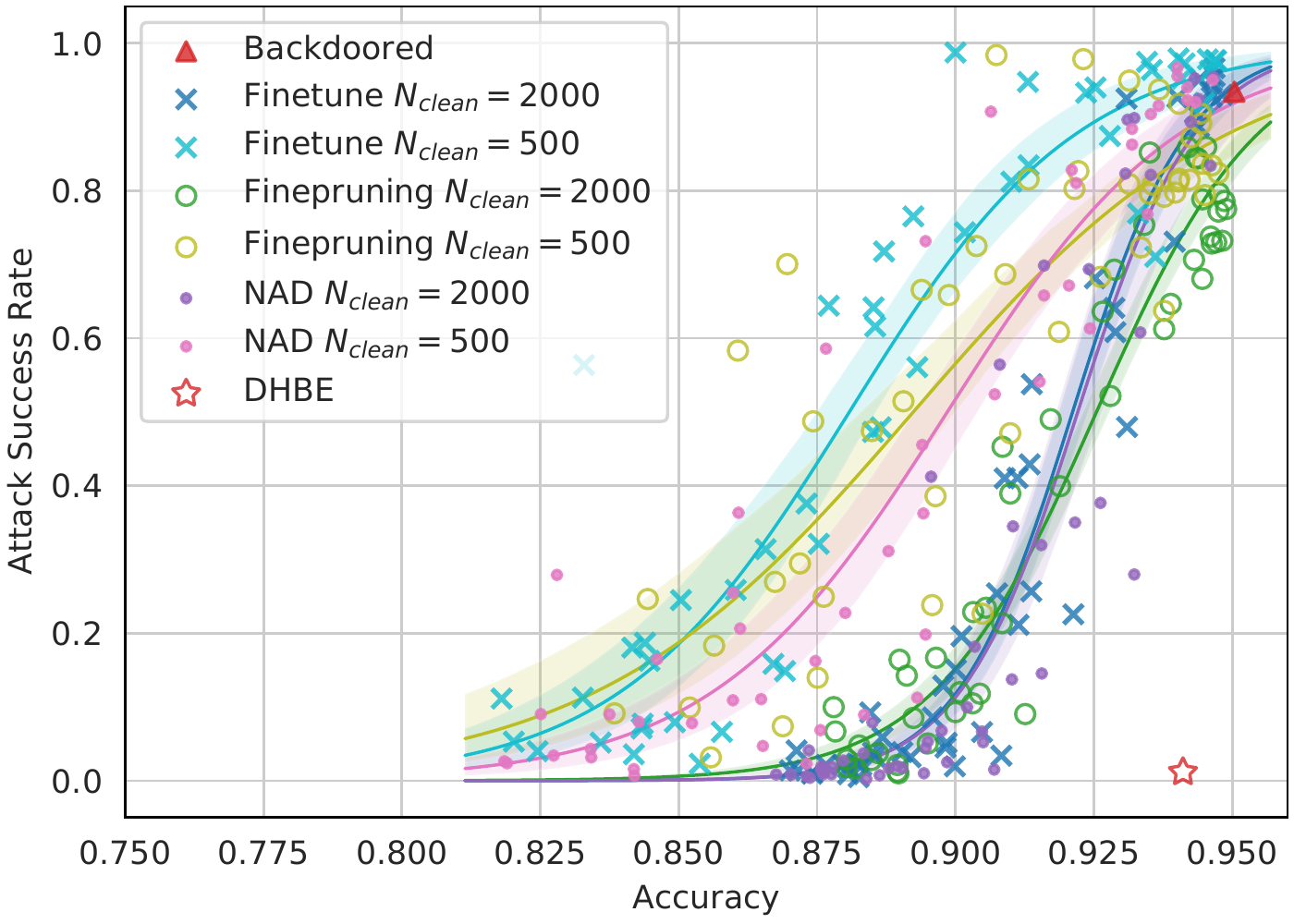}&
    \includegraphics[width=0.30\textwidth]{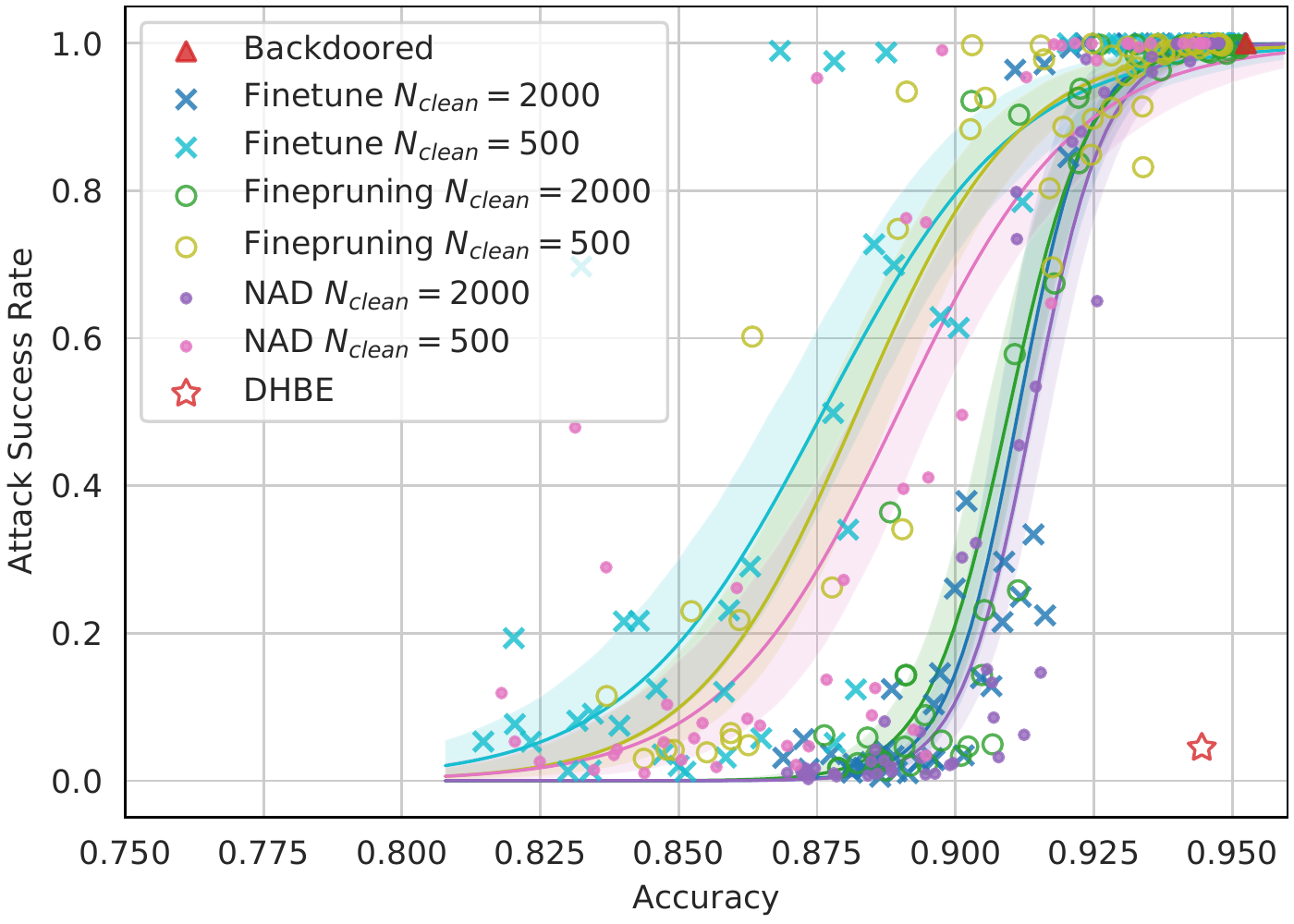}\\
    (a) Clean-label ($2\times 2$)&
    (b) Clean-label ($3\times 3$)&
    (c) Clean-label ($5\times 5$)\\
    \\
    \includegraphics[width=0.30\textwidth]{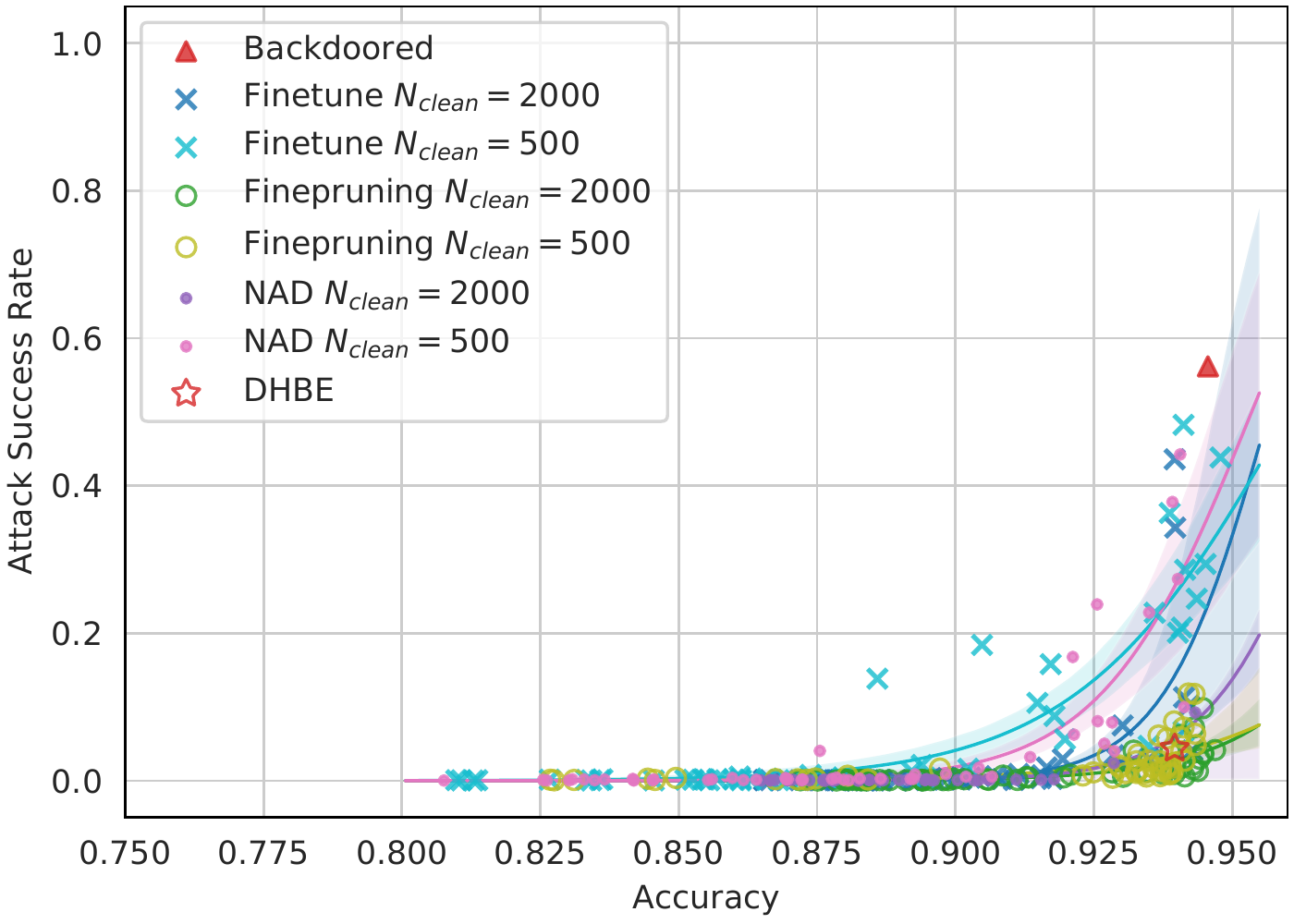}&
    \includegraphics[width=0.30\textwidth]{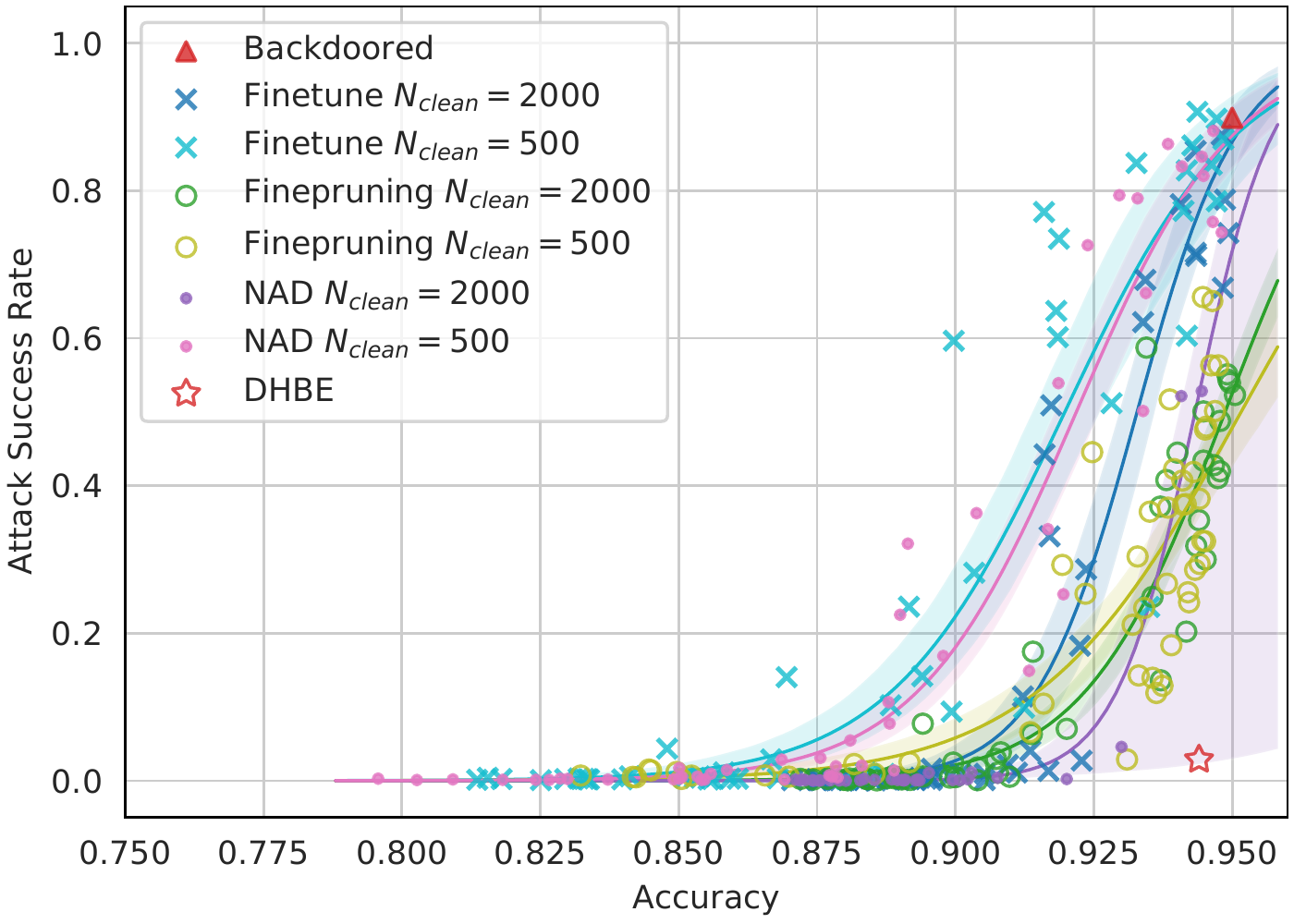}&
    \includegraphics[width=0.30\textwidth]{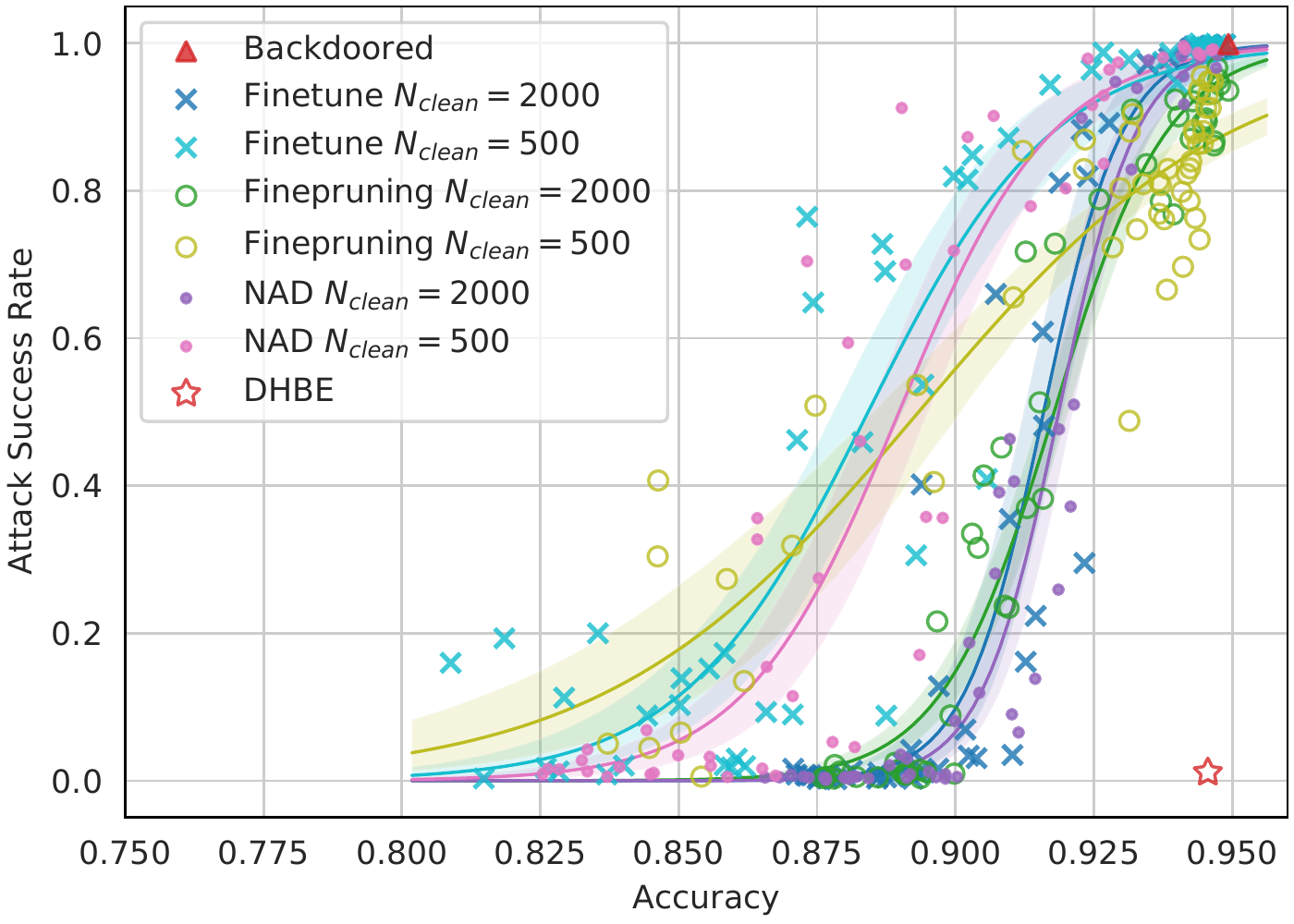}\\
    (d) Trojaning ($2\times 2$)&
    (e) Trojaning ($3\times 3$)&
    (f) Trojaning ($5\times 5$)\\
    \end{tabular}
    \vspace{-3mm}
    \caption{Trade-off curves between accuracy and attack success rate of different blind erasing methods against different triggers. The curves are drawn by adjusting the learning rate of these methods (except DHBE) from 0.002 to 0.02. } 
    \label{fig:trade_off_plot_cifar10_2}
\end{figure*}

\textbf{(b) Model Connectivity Repair (MCR) \cite{zhao2019bridging}.} For MCR, we imitate the open-source code \footnote{\url{https://github.com/IBM/model-sanitization}} and create a Curve net version of employed Resnet-18 (Table \ref{tab:resnet_arch}), where the $conv$, $fc$, $BN$ layers are modified. We set the endpoint of Resnet-18 curve net ($t=0.0$ and $t=1.0$) with the same backdoorded Resnet-18 model, and train the connection path for 200 epochs with SGD optimizer ($lr=0.01$, $momentum=0.9$, $lr decay=0.1$ at epoch 100 and epoch 150). As suggested in \cite{zhao2019bridging}, when evaluating the curve model at its middle point, we first let the model go through the whole test dataset in training mode to correct its running statistics of batch normalization layers, then we fix the batch normalization layers and report its ACC and ASR.


\section{Detailed Configuration of Targeted Methods}

The detailed configuration and training process of targeted methods is summarized as follows:

\textbf{(a) Neural Cleanse \cite{wang2019neural}.} Using the open-source code \footnote{\url{https://github.com/Abhishikta-codes/neural_cleanse}}, we generate a reverse-engineered trigger pattern $\Delta$ and trigger mask $m$ for the target class $t$. Then, $10\%$ of 2000 clean training samples (samples are duplicated if their number is less than 2000) are randomly selected and injected with $\Delta$. Finally, the backdoored model is finetuned using the same set of hyperparameters as the finetuning process in blind methods, except the initial learning rate is $0.001$.

\textbf{(2) Generative Distribution Modeling (GDM) \cite{qiao2019defending}.}
Using the open-source code \footnote{\url{https://github.com/superrrpotato/Defending\%2DNeural\%2DBackdoors\%2Dvia\%2DGenerative\%2DDistribution\%2DModeling}}, we train a generative model for the trigger distribution, using hyperparameters ($\alpha=0.1$, $\beta=0.9$). Then we use the same finetuning process and the same learning rate schedule as Neural Cleanse to finetune backdoored models. For each training sample in each iteration, it is stamped by a randomly sampled trigger with probability 0.1.











\end{document}